\documentclass{midl} % Anonymized submission

\usepackage{float}
\usepackage{caption}
\usepackage{multirow}
\usepackage{hyperref}
\usepackage{booktabs}
\usepackage{graphicx} % DO NOT CHANGE THIS
\usepackage{array}
\usepackage{amsmath,amssymb}

\usepackage{mwe} % to get dummy images
\usepackage{multirow}
\usepackage[figuresright]{rotating}

\jmlryear{2023}
\jmlrworkshop{Full Paper -- MIDL 2023}
\jmlrvolume{-- nnn}
\editors{Accepted for publication at MIDL 2023}

\title[Learning Structured Semantic Consistency]{Toward Unpaired Multi-modal Medical Image Segmentation via Learning Structured Semantic Consistency}

\midlauthor{\Name{Jie Yang\nametag{$^{1}$}} \Email{yangjie5@link.cuhk.edu.cn}\\
\Name{Ye Zhu\nametag{$^{1}$}} 
\Email{zhuye1@cuhk.edu.cn}\\
\Name{Chaoqun Wang\nametag{$^{1}$}} 
\Email{chaoqunwang@link.cuhk.edu.cn}\\
\Name{Zhen Li\nametag{$^{1}$}} 
\Email{lizhen@link.cuhk.edu.cn}\\
\Name{Ruimao Zhang\nametag{$^{1,*}$}} \Email{ruimao.zhang@ieee.org}\\
\addr $^{1}$ Shenzhen Research Institute of Big Data, The Chinese University of Hong Kong, Shenzhen \\
\addr $^{*}$ Corresponding author
}

\begin{document}

\maketitle
\begin{abstract}
% Integrating multi-modal data to promote medical image analysis has received great attention recently. 
% This paper presents a novel scheme to learn the mutual benefits of different modalities to achieve better segmentation results for unpaired multi-modal medical images. Different from previous methods which adopted various data augmentation or module design schemes to accommodate the appearance variance of different modalities, our approach aims to tackle two critical issues of this task from a practical perspective: (1) how to effectively learn the semantic consistencies of various modalities (e.g., CT and MRI), and (2) how to leverage the above consistencies to regularize the network learning while preserving its simplicity. To address (1), our method leverages a carefully designed External Attention Module (EAM) to align semantic class representations and their correlations of different modalities, which enforces segmentation predictions to obey semantic consistency. To solve (2), the proposed EAM is designed as an external plug-and-play one, which can be discarded once the model is optimized. During the testing phase, we only need to maintain the basic segmentation network architecture for cross-modal predictions, which balances the model’s ease of use and simplicity. We have demonstrated the effectiveness of the proposed method on two medical image segmentation scenarios: (1) cardiac structure segmentation, and (2) abdominal multi-organ segmentation. 
Integrating multi-modal data to promote medical image analysis has recently gained great attention. This paper presents a novel scheme to learn the mutual benefits of different modalities to achieve better segmentation results for unpaired multi-modal medical images. Our approach tackles two critical issues of this task from a practical perspective: (1) how to effectively learn the semantic consistencies of various modalities (e.g., CT and MRI), and (2) how to leverage the above consistencies to regularize the network learning while preserving its simplicity. To address (1), we leverage a carefully designed External Attention Module (EAM) to align semantic class representations and their correlations of different modalities. To solve (2), the proposed EAM is designed as an external plug-and-play one, which can be discarded once the model is optimized. We have demonstrated the effectiveness of the proposed method on two medical image segmentation scenarios: (1) cardiac structure segmentation, and (2) abdominal multi-organ segmentation. 
Extensive results show that the proposed method outperforms its counterparts by a wide margin.
\end{abstract}

\section{Introduction}
\label{sec:intro}

% For the assessment of specific diseases, different modalities of imaging, such as computed tomography (CT) and magnetic resonance imaging (MRI), usually provide different information on tissue structure.
% %
% In clinical practice, the data with different physical principles of imaging have always been used together \cite{karim2018algorithms}  to obtain a more comprehensive view of specific organs in disease assessment and treatment planning \cite{cao2017dual}. 
% %
% Although the data from different modalities have dramatic differences in appearance, some similar techniques are always required to aid in the diagnosis, such as quantitative assessment via segmentation on CT/MRI \cite{dou2020unpaired}.
% %
% Previous research primarily focused on developing powerful segmentation models \cite{cao2021swin, gao2021utnet}  for single-modality segmentation, such as CT or MRI. 
% %
% However, Due to the domain shift of different modalities, these well-trained models on one specific modality frequently fail when being deployed to another modality for real-world clinical analysis \cite{chen2020unsupervised}. 
%
%In addition, the separate training scheme is unable to exploit the mutual information from multiple modalities, resulting the sub-optimal accuracy for each modality.
Assessing specific diseases often involves using different imaging modalities, such as CT and MRI, which provide distinct information on tissue structure. In clinical practice, these modalities are combined to achieve a comprehensive understanding of organs for disease assessment and treatment planning \cite{karim2018algorithms, cao2017dual}. Despite the differences in appearance between CT and MRI data, similar techniques like quantitative segmentation are crucial for diagnosis \cite{dou2020unpaired}.
Previous research primarily focused on developing robust segmentation models for single-modality applications \cite{cao2021swin, gao2021utnet}. However, due to domain shifts between modalities, models trained on one modality often fail when applied to another, posing challenges for real-world clinical analysis \cite{chen2020unsupervised}.

In the literature, some recent studies \cite{wang2021transbts,cheng2022fully} have been presented to address the aforementioned issue via joint representation learning from multi-modalities.
%
%They directly align the feature representation of corresponding pixels from different modalities during the training phase.
%
However, this joint representation learning principally necessitates spatial alignment and co-registered sequences within multi-modalities, \textit{e.g.,} multi-sequence MRI (T1, T1c, T2, FLAIR). 
For the unpaired multi-modal data, \textit{e.g.,} CT and MRI, such a scheme is infeasible because of the spatial misalignment.
Recently, Valindria \textit{et al.}~\cite{valindria2018multi} proposed four kinds of dual-steam CNNs to alleviate the negative domain shift between unpaired CT and MRI, where assigning modalities with their specific feature extractors greatly affects the model's parameter efficiency and limits the model's ability to handle more modalities.
Dou \textit{et al.}~\cite{dou2020unpaired} further designed both modality-specific and modality-shared modules to accommodate the appearance variance of different modalities. 
%while extracting the common semantic information.
%
Despite significant efforts to pursue multi-modal medical image segmentation, it still poses some challenges for real-world applications due to the following issues.
%
%
%mining the semantic associations (\textit{e.g.,} the representations of same tissue regions in CT and MRI) and aligning the feature representations from unpaired multi-modal data remain challenging due to the following issues.
%
\textbf{First}, discovering how to fully explore the semantic associations of multiple modalities is critical but very difficult because there is no pixel-to-pixel correspondence in the unpaired input images in practice.
\textbf{Second}, how to discard complex model design and leverage the above semantic associations to regularize the network learning while preserving its simplicity remains intractable nowadays.
%
%\textbf{In addition}, how to maintain reciprocal boosting of multiple modalities and even achieve better segmentation results in many practical situations with limited scarce annotated data for one specific modality has not been well investigated in the literature.
%

To address the above issues, this paper presents a novel method for performing unpaired multi-modal medical image segmentation based on a single Transformer by learning the structured semantic consistency between modalities, \textit{i.e.} the consistencies of semantic class representations and their correlations.
Specifically, the unpaired multi-modal medical images, \textit{e.g.,} CT and MRI, are firstly fed into a shared Transformer backbone to extract multi-scale feature representations.
For each modality, we further introduce a set of modality-specific class embeddings, each of which indicates a global representation of one semantic class.
It is updated during the training phase to learn the specific class representation across the entire dataset.
In practice, these modality-specific class embeddings are learnable and fed into an elaborate External Attention Module (EAM) to interact with the feature maps of the corresponding modal images.
By doing this, the image-specific class embeddings and their correlations of a certain image can be further extracted.
Furthermore, structured semantic consistency across modalities can be achieved gradually by implementing consistency regularizations at the modality-level and image-level respectively.
During the testing phase, we discard all EAMs and only hold a single Transformer for predicting the segmentation results of various modalities.
%

% We conduct extensive experiments on two popular medical image segmentation scenarios with 2D and 3D model configurations.
%
% Experimental results show that our method outperforms the state-of-the-art methods with a large margin, \textit{i.e.} 3.3$\%$ and 2.5$\%$ improvements on the overall mean Dice for two tasks respectively.
%
% Moreover, we evaluate our method with the few-shot segmentation setting on cardiac sub-structure segmentation.
% %
% The proposed method could still achieve competitive performance against state-of-the-art methods, demonstrating the effectiveness of such scheme to deal with scenarios where annotated data for one of the modalities is scarce.
%

In summary, the main contributions of this paper are as follows:
\textbf{(1)} We propose a novel method to learn to segment multi-modal medical images by using a single Transformer backbone.
\textbf{(2)} We introduce a plug-and-play External Attention Module to assist the backbone in discovering the semantic association and learning the structured semantic consistency by using unpaired multi-modal medical images.
\textbf{(3)} We evaluate our method on two different multi-class segmentation tasks with 2D and 3D configurations, showing the effectiveness of our method in various settings. 
\begin{figure*}[ht]
\begin{center}
\includegraphics[width=1\textwidth]{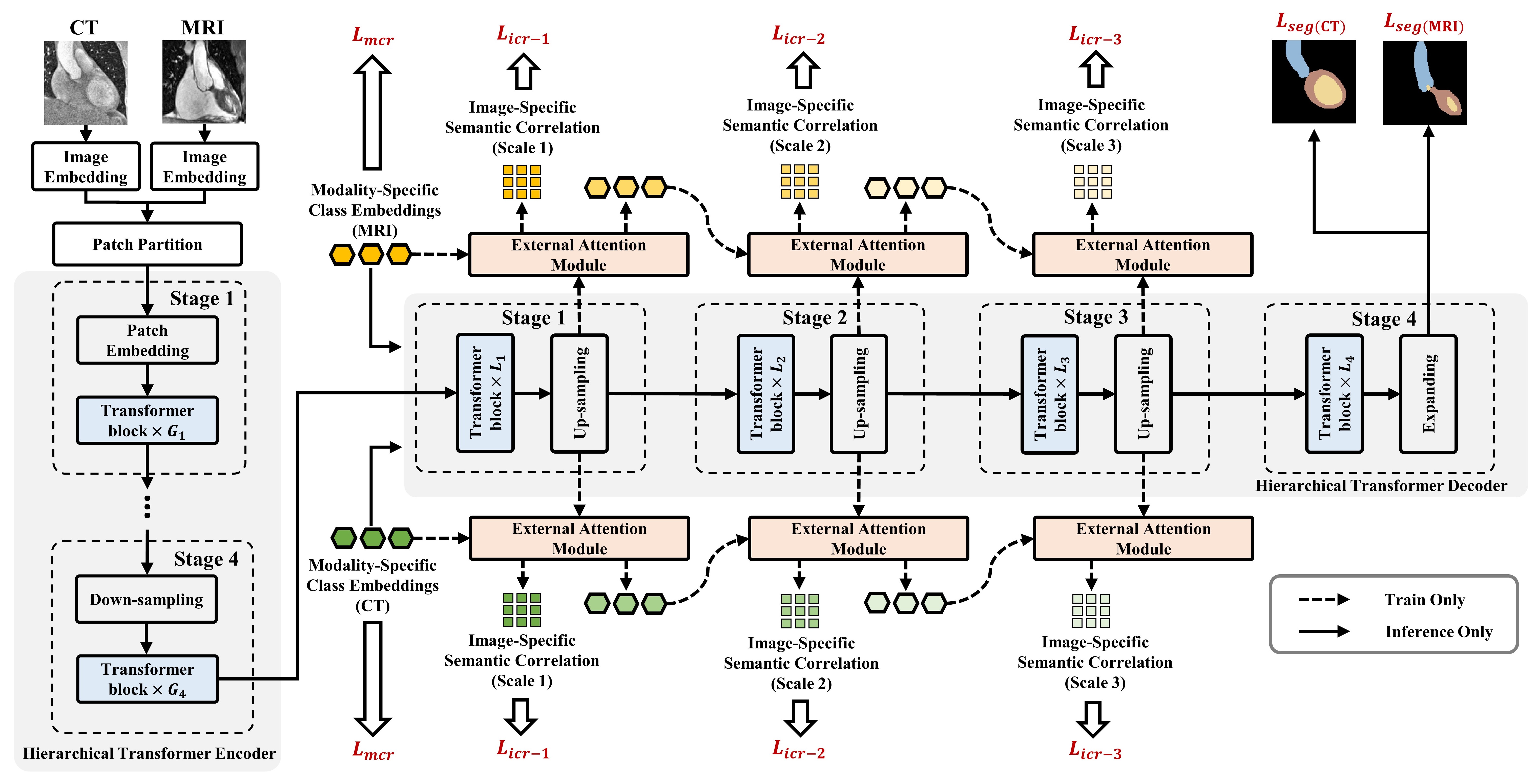}
\end{center}
\vspace{-0.5cm}
\caption{Overview of our proposed unpaired multi-modal medical
image segmentation framework via single Transformer architecture and the proposed EAMs.
%	
% Two unpaired medical images from different modalities are fed into the shared Transformer Encoder and Decoder for feature extraction and dense prediction.
% %
% The \textit{modality-specific class embeddings} for different modalities are introduced to learn modality-specific class representations, while fed into Transformer blocks for realizing the calibration of modality-specific activation (optional operation).
% %
% The newly proposed External Attention Module (EAM) aims to conduct interaction between image features and modality-specific class embeddings, generating \textit{image-specific semantic correlations}. To facilitate the consistency of the two modalities’ representations, the modality-specific class embeddings and image-specific semantic correlations are both aligned between the two modalities at the modality-level and image-level, respectively.
% % for further image-level semantic alignment at multiple scales.
% %
All parts linked by dotted lines can be removed during the inference phase.} 
\label{fig:frmework}
\vspace{-0.7cm}
\end{figure*}

\section{Methodology}
\subsection{Problem Setting and Framework Overview.}
Considering two unpaired medical images $\{\mathbf{X}_{M_1},\mathbf{X}_{M_2} \}$ extracted from different modalities and their corresponding label maps $\{\mathbf{Y}_{M_1},\mathbf{Y}_{M_2} \}$, 
%
%the proposed multi-modal learning framework aims to explore structured semantic information that bridge the modality gap at the training process and to produce a unified segmentation model for multiple modalities' predictions, outperforming the one trained with a single modality.
%
%the proposed multi-modal learning framework aims to explore cross-modality knowledge that explicitly guide the interaction between modalities at training process, to learn a single segmentation model for inference of arbitrary modality that outperforms it trained with one modality alone with a large margin.
%
%
the overall framework of the proposed method is depicted in Fig.~\ref{fig:frmework}. For simplicity, we use the 2D model as an illustration which can be easily extended to the 3D model.
We first feed three consecutive slices as the inputs $\{ \mathbf{X}_{M_1},\mathbf{X}_{M_2} \} \in  \mathbb{R}^{3 \times H \times W} $ into the modality-specific image embedding module, which is simply implemented by two consecutive $1\times1$ convolutional layers, making the resolution and dimension of the inputs unchanged.
The embedded feature maps from the two modalities are then fed into a single Unet-shaped Transformer-based segmentation network that includes encoder and decoder subnetworks for pixel-wise dense prediction. Specifically, the decoders could generate multi-scale features that are combined with the encoders' features, \textit{i.e.} ${2C \times \frac{H}{P} \times \frac{W}{P}}$ at Stage $1$, ${2C \times \frac{H}{2P} \times \frac{W}{2P}}$ at Stage $2$, ${4C \times \frac{H}{4P} \times \frac{W}{4P}}$ Stage $3$, and ${8C \times \frac{H}{8P} \times \frac{W}{8P}}$ at Stage $4$, where $P$ is the patch size in Transformer and is $4$ by default.

To align the modalities during the training phase, we explore two kinds of semantic information, termed \textbf{\textit{modality-specific class embeddings}} and \textbf{\textit{image-specific semantic correlations}}.
The former is a set of learnable vectors for each modality, each of which presents one semantic class, \textit{e.g.,} \texttt{Liver} or \texttt{Spleen}.
It aims to learn the global class representations of each modality. The latter is used to present the inter-class relationships within a specific image.
In practice, a newly designed \textbf{\textit{External Attention Module} (EAM)} is introduced to update the above learnable class embeddings from modality-specific to image-specific, and extract the semantic correlations of a specific image at multiple scales, as shown in Fig.~\ref{fig:frmework}.
% extract the above information via the attentive interaction of learnable proxy embeddings and feature maps of the corresponding modal images.
%
% modality-specific class embeddings of a specific modality are denoted as matrix form $\mathbf{Q} \in  \mathbb{R}^{Z \times 4C}$, where $Z$ indicates the total number of classes and is consistent over different modalities.

% \textbf{For simplicity}, we do not distinguish $\mathbf{Q}$ for different modalities. Thus, the $\mathbf{Q}$ can correspond to the yellow hexagons for CT or the green hexagons for MRI in Fig.~\ref{fig:frmework}.
% %
% We feed $\mathbf{Q}$ into EAM to conduct interaction with the feature maps of the specific image at Scale-1, obtaining image-specific class embeddings $\mathbf{Q}_1 \in \mathbb{R}^{Z \times 2C}$ and semantic correlations $\mathbf{E}_1 \in \mathbb{R}^{Z \times Z}$. 
% %
% To process the multi-scale feature maps, $\mathbf{Q}_\lambda$ is then passed into the EAM at Scale-$(\lambda+1)$ to obtain $\mathbf{Q}_{\lambda+1}$ and $\mathbf{E}_{\lambda+1}$, where $\mathbf{Q}_\lambda$ is adapted to reflect class representations of a specific image within a modality, and $\mathbf{E}_{\lambda}$ presents the image-specific semantic correlations at Scale-$\lambda$. 
% %
% In general, $\mathbf{Q}_{\lambda}\in \mathbb{R}^{Z \times \frac{2C}{\lambda}}, \lambda\in 	\{1,2\}$ and $\mathbf{E}_{\lambda}\in \mathbb{R}^{Z \times Z}, \lambda\in 	\{1,2,3\}$. 
%We will provide a more detailed description of the operations in EAM in the following sections.
% During the training phase, 
We explicitly facilitate the consistency of the two modalities' representations. 
1) We firstly introduce the global consistency regularization $\mathcal{L}_{mcr}$ to minimize the representation distance between modality-specific class embeddings.
It aims to globally align the semantic class representations of two modalities.
%
%so as to transfer structured information from cross-modality to enhance separate category representations.
%
Such consistency will also implicitly affect the pixel-level representation learning of each sample since these modality-specific class embeddings also interact with corresponding images of each modality through the training process.
2) We further align semantic correlations between two modalities at image-level by minimizing $\mathcal{L}_{icr}$ (\textit{i.e.} symmetrical Kullback-Leibler divergence). 
%
%which calculates symmetrical Kullback-Leibler divergence of any two images from different modalities.
%
% Such image-level correlation alignment could further deal with the hard examples when the global category-level representations have already aligned.
%
Such a scheme allows generating many sample pairs to drive semantic correlation alignment, which makes the optimized model more robust to sample variation.
%
% 3) The alignment of image-specific semantic correlations at multiple stages can also overcome the effect of the scales and sizes of the semantic area of different modalities' data.

\subsection{External Attention Module}
%In addition to defining the learnable modality-specific class embeddings to capture the class representations from different modalities, we also propose the External Attention Module (EAM) to conduct the interaction between these modality-specific embeddings and image features to generate class representation of each image, 
%
%and derive semantic correlations of a certain modal image at multiple scales for further semantic alignment. 
%
%
For simplicity, we illustrate the proposed EAM using feature maps $\mathbf{F} \in  \mathbb{R}^{\frac{H}{4P} \times \frac{W}{4P}\times4C}$ at Scale-1 and such mechanism can be applied to other scales as well, as shown in \textbf{Appendix}~\ref{fig:EAM}.

\noindent\textbf{Class Embeddings.} In practice, the learnable modality-specific class embeddings are used to learn the class representations of a specific modality across the entire dataset, while each image should have its own class representations that differ from the global ones due to their appearance variance.
Thus, we employ the Cross-Attention Mechanism (CA) for updating class embeddings by interacting with multi-scale feature maps of a certain image to generate its image-specific class embeddings.
Specifically, we utilize $\mathbf{F}$ to calculate the key and value of CA by linear projection, while the query of CA is calculated employing modality-specific class embeddings $\mathbf{Q} \in  \mathbb{R}^{Z \times 4C}$ as follows:
\begin{gather}
    {\mathbf{q}}=\mathbf{Q}\mathbf{W}_{Q},~{\mathbf{k}}=\mathbf{F}\mathbf{W}_{K},~{\mathbf{v}}=\mathbf{F}\mathbf{W}_{V},\\
    {\rm{CA}}(\mathbf{Q},\mathbf{F})={\rm{softmax}}(\frac{\mathbf{q}\mathbf{k}^\mathsf{T}}{\sqrt{d}})\mathbf{v},
\end{gather}
where $\mathbf{W}_{K}$, $\mathbf{W}_{V}$, $\mathbf{W}_{Q}$ $ \in \mathbb{R} ^ {4C \times 4C^{\prime}}$ are the parameter matrices for linear projection. 
$d$ is the channel dimension of $\mathbf{q}\in \mathbb{R} ^ {Z \times 4C^{\prime}}$ and $\mathbf{k}\in \mathbb{R} ^ {  (\frac{H}{4P} \times \frac{W}{4P} ) \times 4C^{\prime}}$.
The $\rm{softmax(\cdot)}$ denotes the softmax function along the spatial dimension. 
%
%\textit{i.e.} $\frac{H}{4P} \times \frac{W}{4P}$.
%
The $\mathbf{q}\mathbf{k}^\mathsf{T} \in  \mathbb{R}^{Z  \times \frac{H}{4P} \times \frac{W}{4P}}$ indicates the \textit{Semantic-aware Feature Maps} extracted from a single CA head at Scale-1, where $Z$ denotes the total numbers of classes.
The Multi-head Cross Attention (MCA) is the extension with $N$ independent CAs and project their concatenated outputs as follows:
\begin{gather}
    {\rm{MCA}}(\mathbf{Q},\mathbf{F})={\mathbb{C}}(~{\rm{CA}}_{1}(\mathbf{Q},\mathbf{F}),...,{\rm{CA}}_{N}(\mathbf{Q},\mathbf{F})~)~\mathbf{W}_O,
\end{gather}
where $\mathbb{C}$ denotes the concatenation operation. $\mathbf{W}_O$ $ \in \mathbb{R} ^ {4C^{\prime} \times 4C}$ is the learnable parameter matrix, and we have $4C^{\prime} = 4C/N$. 
Here the \textit{Semantic-aware Feature Maps} extracted from multiple attention heads at Scale-1 can be presented as $ \mathbf{A}_{1} \in \mathbb{R}^{Z\times N \times \frac{H}{4P} \times \frac{W}{4P}}$, which can be further adopted to derive image-specific semantic correlations. 
In this way, the $\mathbf{Q}$ can be updated by:
\begin{gather}
    \hat{\mathbf{Q}} = {\rm{MCA}}(~{\rm{Norm}}(\mathbf{Q}),~{\rm{Norm}}(\mathbf{F})~)+\mathbf{Q}, \\
    \tilde{\mathbf{Q}} = {\rm{MLP}}(~{\rm{Norm}}(\hat{\mathbf{Q}})~)+\hat{\mathbf{Q}},
\end{gather}
where $\tilde{\mathbf{Q}}\in  \mathbb{R}^{Z \times 4C}$ reflects image-specific class representations by collecting image-specific semantic information from feature maps of a particular modal image.
In practice, the $1\times1$ convolution operation is further used to reduce the dimension of the above $\tilde{\mathbf{Q}}$ to ${Z \times 2C}$ and obtain image-specific class embeddings $\mathbf{Q}_1 $ for the next scale of updates. 
The operations at the other scales are identical to those described above, and the only difference is that we adopt image-specific class embeddings (\textit{e.g.} $\mathbf{Q}_1 $ or $\mathbf{Q}_2 $) to replace $\mathbf{Q}$ in the above equations.

\noindent\textbf{Semantic Correlations.} 
The semantic correlations reflect the inter-class representation similarities.
In practice, we propose a three-step process to extract the semantic correlation matrix in a specific image.
Again, we explain the corresponding operations at Scale-1 as an illustration.
Given the modality-specific class embeddings $\mathbf{Q}$ and 
multi-head semantic-aware feature maps $\mathbf{A}_{1}$, the semantic correlations $\mathbf{E}_1$ can be calculated via semantic filtering, semantic re-weighting, and semantic aggregation, respectively.
The operations at the other scales are the same, the only difference is that we replace $\mathbf{Q}$ with image-specific class embeddings $\mathbf{Q}_1 $ and $\mathbf{Q}_2 $ to calculate $\mathbf{E}_2$ and $\mathbf{E}_3$.

\textbf{\textit{1) Semantic Filtering.}} Given a particular semantic class embeddings in $\mathbf{Q}$, 
the purpose of such a step is to calculate its relevance to all semantic classes at the token level.
Since  $\mathbf{A}_{1}  \in \mathbb{R}^{ Z \times N \times (\frac{H}{4P} \times \frac{W}{4P}) }$, the dimension of tokens is $ Z\times N$.
We can divide the features of tokens into $Z$ groups, each of which is corresponding to a particular semantic class.
Thus we have $\mathbf{A}_{1} = \{\mathbf{A}_1^j\}_{j=1}^Z$ and $ \mathbf{A}_{1}^j \in \mathbb{R}^{N \times \frac{H}{4P} \times \frac{W}{4P}}$.
Similarly, we can rewrite $\mathbf{Q}$ as $\{\mathbf{Q}^i\}_{i=1}^Z$ and $\mathbf{Q}^{i} \in \mathbb{R}^{4C}$.
For $i$-th class, we generate the semantic kernel as $\mathbf{K}^{i}=\mathbf{Q}^{i}\mathbf{W}^{i}$, where $\mathbf{W}^{i}\in \mathbb{R}^{4C\times N}$ is the parameter matrix that is corresponding to the $i$-th class.
To calculate the similarity between $i$-th and $j$-th classes at token level, we can directly reshape $\mathbf{K}^{i}$ into $\mathbb{R}^{N\times 1 \times1}$ to perform filtering on $\mathbf{A}_{1}^j$ as follows,
\begin{equation}
    \mathbf{S}_1^{ij} = \mathbb{F}( \mathbf{K}^{i}, \mathbf{A}_1^j ),
\end{equation}
where the function $\mathbb{F(\cdot,\cdot)}$ denotes the convolutional operation, and $\mathbf{S}_1^{ij}\in \mathbb{R}^{\frac{H}{4P} \times \frac{W}{4P}}$ is a similarity map of $i$-th class for the $j$-th class,
where tokens with higher response scores indicate the higher correlation to $i$-th class representation in $\mathbf{Q}$.
That is reasonable in practice. 
For example, the \texttt{left kidneys} and \texttt{right kidneys} should have similar structural representations due to their similar appearance, shape, and size. 
Thus, the class filter of \texttt{left kidneys}  should also have the highest response to the region of the \texttt{right kidneys}.

%There usually exists a higher probability of inter-class similarity than in other organs. 
%
%Thus, the class filter of \texttt{left kidneys}  should have the highest response to the region of the \texttt{left kidneys}, followed by the second-highest response to the region of the \texttt{right kidneys} and a lower response to the other organs.

%
\textbf{\textit{2) Semantic Re-weighting.}} Given similarity maps between $i$-th class representation in $\mathbf{Q}$ and all groups' feature maps of $\mathbf{A}_{1}$, 
then we have $ \mathbf{S}_1^i = \{\mathbf{S}_1^{i1}, ..., \mathbf{S}_1^{ij}, ..., \mathbf{S}_1^{iZ}  \} \in \mathbb{R}^{Z \times (\frac{H}{4P} \times \frac{W}{4P})}$.
By conducting the softmax operation on each spatial position of $ \mathbf{S}_1^i$, 
each element in $\mathbf{A}_{1}$ is  weighted by the gating function as follows,

\begin{equation}
    \mathbf{B}_1^i=\mathbf{A}_{1}\odot \mathbb{B}(\sigma(\mathbf{S}_1^i)),
\end{equation}
where $\sigma(\cdot)$ is softmax operation, $\mathbb{B}$ is the broadcast operation to extend the dimension of input to $ Z\times N \times \frac{H}{4P} \times \frac{W}{4P}$, 
and $\odot$ denotes the element-wise multiplication.
In this way, we obtain $\mathbf{B}_1^i=\{\mathbf{B}_1^{i1}, ..., \mathbf{B}_1^{ij}, ..., \mathbf{B}_1^{iZ}  \}  \in \mathbb{R}^{Z \times N \times (\frac{H}{4P} \times \frac{W}{4P})} $, 
where $\mathbf{B}_1^{ij}$ denotes the correlation map between $i$-th class representation in $\mathbf{Q}$ and $j$-th class feature maps.

\textbf{\textit{3) Semantic Aggregation.}}
To generate the final correlation map, we conduct the normalized summation on $\mathbf{B}_1^i$ along the last three dimensions to realize semantic aggregation,
\begin{equation}
    \mathbf{E}_1^i=\frac{\sum_{(N ,\frac{H}{4P}, \frac{W}{4P})}\mathbf{B}_1^i}{\sum_{(\frac{H}{4P}, \frac{W}{4P})}{ \sigma(\mathbf{S}_1^{i}) }}
\end{equation}
where $\mathbf{E}_{1}^{i} \in \mathbb{R}^Z$ is the normalized correlation vector, and each element in $\mathbf{E}_{1}^{i}$ presents the relevance between the $i$-th class representation in $\mathbf{Q}$ and one specific class representation of the input image.
Finally, image-specific semantic correlations at Scale-1 can be presented as $\mathbf{E}_{1}=\{\mathbf{E}_1^1, ..., \mathbf{E}_1^i, ..., \mathbf{E}_1^Z\}\in \mathbb{R}^{Z \times Z}$.
The operations of EAM at other scales are the same as Scale-1's, but with different input feature dimensions.
In our scheme, the proposed EAM outputs the above semantic correlations at Scale-1 for each modality, \textit{i.e.} denoted by $\mathbf{E}_{1:M_1}$ and $\mathbf{E}_{1:M_2}$ for CT and MRI, which can be continuously updated during the training process. 
We motivate to dynamically align the semantic correlations of images from different modalities in the training phase since we intuitively assume that the inter-class relationships shown in CT are still valid in MRI.

%We advocate for dynamically aligning the semantic correlations of images from different modalities in Objective Functions~\ref{sec:loss}, as we intuitively assume that the inter-class relationships shown in CT are still valid in MRI. For example, the inter-class relationship between the \texttt{left kidneys} and \texttt{right kidneys} in abdominal images should be similar.

%%%%%%%%%%%%%  附件材料
\iffalse
It is worth noting that 
%
since there is no spatial alignment between two cross-modal images (\textit{e.g.}  CT and MRI),  there may have a huge difference in the distribution of semantic classes between these images and may obtain a highly imbalanced response for the specific class representation in $\mathbf{Q}$.
%
When aligning the image-specific semantic correlations between two cross-modal images in such a case,
%
it may raise unstable in the training process.
%
Here we calculate normalized summation by dividing by $\sum_{(\frac{H}{4P}, \frac{W}{4P})}{ \sigma(\mathbf{S}_1^{i}) }$ to avoid such a situation.
%
\fi

\subsection{Objective Functions}
\label{sec:loss}
\textbf {Auxiliary Prediction Loss.} 
We introduce an auxiliary loss to supervise the semantic prediction of each pixel on multi-scale semantic-aware feature maps by using cross-entropy (CE) and Dice loss (DSC),
%
%Concretely, we employ convolution layer over semantic-aware feature maps at each scale to predict the segmentation labels,
%
%which is supervised by the auxiliary ground-truth at corresponding scales using cross-entropy (CE) and Dice loss (DSC). 
%
\begin{equation}
\begin{split}
        \mathcal{L}_{\rm{aux}}=\sum_{\lambda}  \Big[  {\rm{CE}}(\mathbb{F}_{1\times 1}({\mathbf {A}}_{\lambda}),\mathbf{Y}_{\lambda})+ {\rm{DSC}}(\mathbb{F}_{1\times 1}({\mathbf {A}}_{\lambda}),\mathbf{Y}_{\lambda}) \Big],
\end{split}
\label{eq:auxiliary}
\end{equation}
where $\lambda \in 	\{1,2,3\}$ represents three scales, and $\mathbb{F}_{1\times 1}(\cdot)$ denotes the function with a $1\times1$ convolution operation to reduce the dimension of multi-head.

\noindent\textbf {Modality-level Consistency Regularization.}
We introduce consistency regularization to globally align the class representations of two modalities.
Let $\mathbf{Q}_{M_1}$ and $\mathbf{Q}_{M_2} \in  \mathbb{R}^{Z \times 4C}$ denote modality-specific class embeddings of two modalities. 
Then the modality-level consistency regularization can be presented as follows:
\begin{equation}
        \mathcal{L}_{\rm{mcr}}= \sum_{i=1}^Z (1-\frac{\mathbf{Q}_{M_1}^{i~\mathsf{T}} \mathbf{Q}_{M_2}^i}{\Vert \mathbf{Q}_{M_1}^i \Vert \cdot \Vert \mathbf{Q}_{M_2}^i \Vert}),
\label{eq:mcr}
\end{equation}
where $Z$ denotes the total number of semantic classes.

\noindent\textbf {Image-level Consistency Regularization.}
We further utilize the symmetrical Kullback-Leibler (KL) divergence to locally align image-specific semantic correlations of each modality.
For two modalities $M_1$ and $M_2$,
let $\mathbf{E}_{\lambda:M_1}^{i}$ and $\mathbf{E}_{\lambda:M_2}^{i}$ denote  correlation vectors corresponding to class $i$-th at Scale-$\lambda$. 
The image-level consistency regularization can be presented as follows:
\begin{equation}
\begin{split}
    \mathcal{L}_{\rm{icr}}=& \sum_{\lambda} \sum_{i=1}^Z \Big[~ {\mathcal{D}_{\rm{KL}}}(\sigma(\mathbf{E}_{\lambda:M_1}^{i}/\tau)~||~\sigma(\mathbf{E}_{\lambda:M_2}^{i}/\tau)) \\
    & +{\mathcal{D}_{\rm{KL}}}(\sigma(\mathbf{E}_{\lambda:M_2}^{i}/\tau)~||~\sigma(\mathbf{E}_{\lambda:M_1}^{i}/\tau))~ \Big],
\end{split}
\label{eq:icr}
\end{equation}
where $\mathcal{D}_{\rm{KL}}(\cdot||\cdot)$ denotes the relative entropy. 
The $\sigma(\cdot)$ denotes the softmax operation along the class dimension. 
$\tau$ is a temperature hyper-parameter to control the softness. 

%of the related probability distributions.
%

\noindent\textbf {Overall Objective Function.}
The overall objective function of the proposed method can be presented as follows:
\begin{equation}
        \mathcal{L}=\mathcal{L}_{\rm{seg}}^{M_1}+\mathcal{L}_{\rm{seg}}^{M_2}+\alpha( \mathcal{L}_{\rm{aux}}^{M_1}+\mathcal{L}_{\rm{aux}}^{M_2})+\beta \mathcal{L}_{\rm{mcr}}+ \gamma \mathcal{L}_{\rm{icr}}, 
\end{equation}
where $\mathcal{L}_{\rm{seg}}^{M_1}$ and $\mathcal{L}_{\rm{seg}}^{M_2}$ denote the segmentation losses for modality $M_1$ and $M_2$ respectively.
%
%which are combined with their Dice losses and cross-entropy losses. 
%
Similarly, the $\mathcal{L}_{\rm{aux}}^{M_1}$ and $\mathcal{L}_{\rm{aux}}^{M_2}$ indicate auxiliary prediction losses for two modalities.
%
%The hyper-parameters $\alpha, \beta, \gamma$ are adopted to scale the contribution of each loss to the overall objective function. 
%
% And we empirically set the hyper-parameters $\alpha, \beta, \gamma$ as $0.5$, $0.5$, and $0.5$.

%to balance them and the main segmentation losses. 

\section{Experiments}
We evaluate the performance of our method on the two multi-modality segmentation tasks, \textit{i.e.} Cardiac Substructure Segmentation and Abdominal Multi-organ Segmentation, under 2D and 3D model configurations respectively. For fairness, we provide four experimental settings: (1) \text{Backbone} that is separately trained with single modality; (2) \text{Baseline} that adds auxiliary prediction loss $\mathcal{L}_{\rm{aux}}$ based on \text{Backbone}; (3) \text{Joint Training} that shares the entire backbone model to jointly train multiple modalities; (4) \text{Ours (w/o CR)} that introduces the modality-aware channel-wise multiplication mechanism in each transformer block of the shared encoder and decoder, as illustrated in \textbf{Appendix}~\ref{sec:transformer}; (5) \text{Ours (w/ CR)} that is our full cross-modal learning strategy by adding two types of consistency terms $\mathcal{L}_{\rm{mcr}}$, $\mathcal{L}_{\rm{icr}}$. We assess segmentation performance using the Volume Dice Coefficient (Dice, $\%$) and Average Symmetric Surface
Distance (ASD, $mm$) metrics.
%
% For fairness,
% We take into account the four experimental settings that conduct the fixed 2D or 3D Transformer and hyper-parameters for a fair comparison of the different settings. we give the single-domain result We compare our method with the state-of-the-arts and the Backbone and Baseline model with the best performance highlighted in \textbf{bold}., and 
% %

% The improvement between our method and baseline model is highlighted in \textcolor[RGB]{254,76,97}{red}.

% 1) \textbf {Backbone:} training a backbone separately by using each modality.

% 2) \textbf {Baseline:} training a separate backbone while adding auxiliary prediction loss $\mathcal{L}_{\rm{aux}}$ for each modality.

% 3) \textbf {Joint Training:}  training a joint Transformer with shared encoder and decoder and implementing the modality-aware channel-wise multiplication mechanism in each transformer block of the shared encoder and decode.

% 4) \textbf {Ours (Joint Training + CR):} our full multi-modal learning strategy, utilizing \textit{Joint Training} Transformer and two types of consistency terms $\mathcal{L}_{\rm{mcr}}$, $\mathcal{L}_{\rm{icr}}$.

\subsection{Cardiac Substructure Segmentation}

\textbf{2D Configuration.}
We employ the Multi-Modality Whole Heart Segmentation Challenge 2017 dataset \cite{zhuang2019evaluation} to perform multi-class cardiac structure segmentation. We adopt a 2D U-shaped Transformer named Swin-Unet \cite{cao2021swin} as the backbone. Please refer to the \textbf{Appendix}~\ref{sec:imple} for more details about the dataset and network.
%
%: \texttt{left ventricle myocardium} (LVM), \texttt{left atrium blood cavity} (LAC), \texttt{left ventricle blood cavity} (LVC), and \texttt{ascending aorta} (AA). 
%
% The dataset is composed of unpaired $20$ CT and $20$ MRI scans collected from various patients and sites.
% %
% % We intend to train the segmentation network to recognize four organs for each modality: \texttt{left ventricle myocardium} (LVM), \texttt{left atrium blood cavity} (LAC), \texttt{left ventricle blood cavity} (LVC), and \texttt{ascending aorta} (AA). 
% Since the UMMKD approach \cite{dou2020unpaired} achieves state-of-the-art performance, we take the same partitioning (\textit{i.e.} training, validation and test) and pre-processing of the dataset as the UMMKD reported for a fair comparison. 
%
% Specifically, we align with them to split each modality into $70\%$ for training, $10\%$ for validation, and $20\%$ for testing. Before feeding into the network, we first resample both modalities to isotropic 1$\rm{mm}^3$ and crop their central heart region by a 3D bounding box which has a fixed coronal plane size of $256 \times 256$. For each 3D cropped image, we remove the top $2\% $ of its intensity histogram to reduce artifacts, which is then normalized to zero mean and unit variance.

\begin{table*}
\centering
\resizebox{\textwidth}{!}{
\begin{tabular}{rl|ccccl|ccccl|l} 
\hline
\toprule [1.5 pt]
\multicolumn{2}{c|}{\multirow{2}{*}{Methods}} & \multicolumn{5}{c|}{Cardiac CT} &
\multicolumn{5}{c|}{Cardiac MRI} & \multirow{2}{*}{Overall Mean} \\
\cline{3-12} 
   &  &LVM & LAC & LVC & AA & Mean &  LVM & LAC & LVC & AA & Mean & \\ \specialrule{0.05em}{0pt}{3pt}
\multicolumn{13}{c}{Dice Coefficient (avg.$\pm$ std., $\%$) $\uparrow$}  \\ \specialrule{0.05em}{3pt}{0pt}
Payer \textit{et al.}\!\!\!&\!\!\!\scriptsize[\textcolor[RGB]{52 152 219}{MMWHS18}] & 87.2$\pm$3.9& 92.4$\pm$3.6 & 92.4$\pm$3.3& 91.1$\pm$18.4& 90.8& 75.2$\pm$12.1& 81.1$\pm$13.8 &87.7$\pm$7.7 &76.6$\pm$13.8 &80.2 &85.5 \\ 
UMMKD\!\!\!&\!\!\!\scriptsize[\textcolor[RGB]{52 152 219}{TMI20}]& 88.5$\pm$3.1&  91.5$\pm$3.1& 93.1$\pm$2.1& 93.6$\pm$4.3& 91.7& 80.8$\pm$3.0& 86.5$\pm$6.5 &93.6$\pm$1.8 &83.1$\pm$5.8 &86.0 &88.8 \\ 
\multicolumn{2}{c|}{Backbone} & 90.0$\pm$3.2&  92.5$\pm$2.9& 92.6$\pm$3.0 & 87.4$\pm$3.8& 90.6& 79.9$\pm$4.6 & 85.3$\pm$3.9 &92.0$\pm$2.7 &84.9$\pm$2.9 &85.5& 88.1 \\ 
\multicolumn{2}{c|}{Baseline} & 90.6$\pm$2.8&  92.6$\pm$2.8& 93.2$\pm$2.5& 88.9$\pm$3.4& 91.3 & 80.9$\pm$4.0& 86.3$\pm$3.8 &92.9$\pm$2.3 &85.8$\pm$3.5 &86.5 & 88.9 \\ \hline  
\multicolumn{2}{c|}{Joint Training} & 89.1$\pm$2.8&  93.0$\pm$2.7& 92.8$\pm$3.3& 91.2$\pm$2.6& 91.5 (\textcolor[RGB]{254,76,97}{+0.2}) & 80.2$\pm$3.9& 86.5$\pm$4.5 &92.0$\pm$3.0 &86.1$\pm$3.8 &86.2 (\textcolor[RGB]{0 205 102}{-0.3})  & 88.9 (+0.0) \\ 
\multicolumn{2}{c|}{Our(w/o CR)} & 90.0$\pm$2.3&  93.8$\pm$2.1& 93.4$\pm$2.4& 94.0$\pm$2.0& 92.8 (\textcolor[RGB]{254,76,97}{+1.5}) & 81.0$\pm$3.1& 87.4$\pm$3.6 &93.5$\pm$2.1 &87.8$\pm$3.0 &87.4 (\textcolor[RGB]{254,76,97}{+0.9})  & 90.1 (\textcolor[RGB]{254,76,97}{+1.2}) \\ 

\multicolumn{2}{c|}{\textbf{Ours(w/ CR)}} & \textbf{90.9$\pm$2.0} &  \textbf{94.8$\pm$1.6} & \textbf{94.5$\pm$2.1} & \textbf{95.9$\pm$1.4} & \textbf{94.0} (\textcolor[RGB]{254,76,97}{+2.7}) & \textbf{81.6$\pm$2.5} & \textbf{89.6$\pm$3.3} & \textbf{94.4$\pm$1.3} & \textbf{89.2$\pm$2.8} & \textbf{88.7} (\textcolor[RGB]{254,76,97}{+2.2}) & \textbf{91.4} (\textcolor[RGB]{254,76,97}{+2.5}) \\ \specialrule{0.05em}{0pt}{3pt}
\multicolumn{13}{c}{Average Symmetric Surface
Distance (avg.$\pm$ std., $mm$) $\downarrow$}  \\ \specialrule{0.05em}{3pt}{0pt} Payer \textit{et al.}\!\!\!&\!\!\!\scriptsize[\textcolor[RGB]{52 152 219}{MMWHS18}] &  -& -& -& -& -& - &- &- &- & - \\
UMMKD\!\!\!&\!\!\!\scriptsize[\textcolor[RGB]{52 152 219}{TMI20}] &  -& -& -& -&-& -&- &- &- &- \\ 
\multicolumn{2}{c|}{Backbone}& 1.67$\pm$0.46&  1.95$\pm$0.54& 1.43$\pm$0.47& 1.51$\pm$0.41& 1.64& 2.12$\pm$1.57 &1.74$\pm$0.85 & 1.41$\pm$0.81 & 3.74$\pm$1.68& 2.25 & 1.95  \\ 
\multicolumn{2}{c|}{Baseline} & 1.49$\pm$0.33&  1.84$\pm$0.44& 1.38$\pm$0.35& 1.46$\pm$0.28& 1.54& 1.71$\pm$1.43& 1.37$\pm$0.64 &1.46$\pm$0.89 &2.69$\pm$1.27 &1.86 & 1.70 \\ \hline
\multicolumn{2}{c|}{Joint Training} & 1.58$\pm$0.35&  1.70$\pm$0.44& 1.39$\pm$0.35& 1.33$\pm$0.38& 1.50 (\textcolor[RGB]{254,76,97}{-0.04}) & 1.87$\pm$0.92& 1.47$\pm$0.40 &1.42$\pm$0.55 &3.13$\pm$1.41 &1.97 (\textcolor[RGB]{0 205 102}{+0.11}) & 1.74 (\textcolor[RGB]{0 205 102}{+0.04}) \\ 
\multicolumn{2}{c|}{Our(w/o CR)} & 1.34$\pm$0.31&  1.63$\pm$0.46& 1.32$\pm$0.27& 1.10$\pm$0.29&1.35 (\textcolor[RGB]{254,76,97}{-0.19}) & 1.84$\pm$0.81& \textbf{1.22$\pm$0.53} &1.39$\pm$0.58 &2.05$\pm$1.10 &1.63 (\textcolor[RGB]{254,76,97}{-0.23}) & 1.49 (\textcolor[RGB]{254,76,97}{-0.21}) \\ 

\multicolumn{2}{c|}{\textbf{Ours(w/ CR)}} & \textbf{1.31}$\pm$0.27&  \textbf{1.49$\pm$0.38}& \textbf{1.22$\pm$0.27}& \textbf{1.00$\pm$0.24}& \textbf{1.26} (\textcolor[RGB]{254,76,97}{-0.28})& \textbf{1.55$\pm$0.78}& 1.24$\pm$0.34 &\textbf{1.27$\pm$0.32} &\textbf{2.01$\pm$0.95} &\textbf{1.52} (\textcolor[RGB]{254,76,97}{-0.34}) & \textbf{1.39} (\textcolor[RGB]{254,76,97}{-0.31}) \\ 
\hline

 \end{tabular}}
 
  \caption{The performance of cardiac substructure segmentation by using 2D Transformer.
  %by \textcolor[RGB]{0 205 102}{green}.
  }
 \label{tbl:cardiac}
 \vspace{-0.7cm}
\end{table*}

\textbf{Main Results.} 
% 等一下放实验表格和相关图
Based on the $2$D Backbone, we extend our \text{Baseline} model by the auxiliary prediction loss $\mathcal{L}_{\rm{aux}}$ in Eqn.~\ref{eq:auxiliary},
which achieves average segmentation Dice of $91.3\%$ on CT and $86.5\%$ on MRI and outperforms the \text{Backbone} model, as well as the MICCAI-MMWHS challenge winner Payer \textit{et al.} \cite{payer2017multi} that also deploys single modality training. This demonstrates that similar to a form of deep supervision, calibrating multi-scale semantic-aware feature maps improves final segmentation performance noticeably.
% 如果不放网络结构细图，这部分需要删改

For multi-modality training, we share the entire backbone model (\textit{i.e.}, encoder, decoder, and prediction head) for multi-modality training, denoted as \text{Joint Training}. However, there is a decrease in segmentation results, \textit{i.e.} Dice of $91.5\%$ on CT and $86.2\%$ on MRI. 
This suggests that in such a situation, modality discrepancy has a significant impact on learned feature representations.
We then introduce a modality-aware channel-wise multiplication mechanism into each Transformer block of shared encoder and decoder based on the \text{Joint Training} model, as in \textbf{Appendix}~\ref{sec:transformer}. This training scheme denoted as \text{Ours (w/o CR)} further improves segmentation results to $92.8\%$ on CT and $87.4\%$ on MRI, demonstrating the efficiency of modality-specific activation calibration. 
Finally,
% \textit{i.e.} $\mathcal{L}_{\rm{mcr}}$ and $\mathcal{L}_{\rm{icr}}$, both of the multi-modal Transformer architectures are boosted.
%
\text{Ours (w/o CR)} with $\mathcal{L}_{\rm{mcr}}$ and $\mathcal{L}_{\rm{icr}}$ marked as \text{Ours (w/ CR)} in Table~\ref{tbl:cardiac} achieves an overall Dice of $91.4\%$ (\textit{i.e.} the average of $94.0\%$ on CT and $88.7\%$ on MRI). 
When compared to the current state-of-the-art multi-modal approach UMMKD \cite{dou2020unpaired}, our segmentation result has a $2.6\%$ promotion on overall mean Dice. 
In addition, our method achieves the lowest overall mean ASD (\textit{i.e.} 1.39mm) among all compared approaches. 
We also present a visual representation of the segmentation results for quantitative comparison in \textbf{Appendix}~\ref{fig:cardiac_abdominal}. 

% \begin{figure*}[ht]
% 	\begin{center}
% 		\includegraphics[width=0.90\linewidth]{figure/cardiac_vi.jpg}
% 	\end{center}
% 	\vspace{-0.3cm}
% 	\caption{Visualization results on cardiac segmentation task and abdominal multi-organ task. \textbf{Left:} AA, LAC, LVC and LVM. \textbf{Right:} Spleen, R-kdy, L-kdy, and Liver. The corresponding colormaps are \textbf{\textcolor[RGB]{238,37,36}{red}}, \textbf{\textcolor[RGB]{145, 193,  62}{green}}, \textbf{\textcolor[RGB]{29, 162, 220}{blue}} and \textbf{\textcolor[RGB]{254, 232,  81}{yellow}}, respectively.} 
% \label{fig:cardiac}
%  \vspace{-0.5cm}
% \end{figure*}

\begin{table*}

\centering
\resizebox{\textwidth}{!}{
\begin{tabular}{rl|ccccl|ccccl|l} 
\toprule [1.5 pt]

\multicolumn{2}{c|}{\multirow{2}{*}{Methods}} & \multicolumn{5}{c|}{Abdominal CT} &
\multicolumn{5}{c|}{Abdominal MRI} & \multirow{2}{*}{Overall Mean}\\
\cline{3-12} 
   & & Liver & Spleen & R-kdy & L-kdy & Mean &  Liver & Spleen & R-kdy & L-kdy & Mean & \\ \specialrule{0.05em}{0pt}{3pt}
\multicolumn{13}{c}{Dice Coefficient (avg.$\pm$ std., $\%$) $\uparrow$}  \\ \specialrule{0.05em}{3pt}{0pt}
UMMKD\!\!\!&\!\!\!\scriptsize[\textcolor[RGB]{52 152 219}{TMI20}] & 92.7$\pm$1.8&  93.7$\pm$1.7& \textbf{94.0$\pm$0.7}& 89.5$\pm$3.9& 92.4& 90.3$\pm$2.8&87.4$\pm$1.1 &91.0$\pm$1.5 &88.3$\pm$1.7 &89.3 & 90.8 \\ 
\multicolumn{2}{c|}{Backbone} & 93.2$\pm$2.8 & 90.9$\pm$2.4&  86.9$\pm$2.7 & 87.5$\pm$3.8 & 89.5& 91.7$\pm$4.0 & 87.2$\pm$3.2 &90.9$\pm$2.7 &90.6$\pm$3.3 &90.1 & 89.9 \\  
\multicolumn{2}{c|}{Baseline} & 93.6$\pm$2.2&  92.1$\pm$2.6 & 87.9$\pm$1.8 & 87.7$\pm$3.6 & 90.3 & 92.9$\pm$3.3& 87.8$\pm$2.9 &92.0$\pm$2.5 &91.4$\pm$3.1 &91.0 & 90.7 \\  \hline
\multicolumn{2}{c|}{Joint Training} & 94.0$\pm$2.6 &  92.5$\pm$2.2 & 87.8$\pm$2.9 & 87.9$\pm$3.4 & 90.6 (\textcolor[RGB]{254,76,97}{+0.3})& 92.6$\pm$3.4 & 87.3$\pm$3.1 & 91.2$\pm$1.9 &90.8$\pm$3.7 &90.5 (\textcolor[RGB]{0 205 102}{-0.5})& 90.5 (\textcolor[RGB]{0 205 102}{-0.2}) \\ 
\multicolumn{2}{c|}{Ours(w/o CR)} & 94.6$\pm$2.2 &  93.3$\pm$1.8 & 88.9$\pm$2.3 & 88.7$\pm$3.5& 91.4 (\textcolor[RGB]{254,76,97}{+1.1})& 93.8$\pm$2.3 & 88.5$\pm$2.7 &92.7$\pm$1.5 &91.5$\pm$3.1 &91.6 (\textcolor[RGB]{254,76,97}{+0.6}) & 91.5 (\textcolor[RGB]{254,76,97}{+0.8}) \\ 
\multicolumn{2}{c|}{\textbf {Ours(w CR)}} & \textbf{95.8$\pm$1.4} &  \textbf{94.9$\pm$1.3} & 92.3$\pm$0.9 & \textbf{91.8$\pm$2.2} & \textbf{93.7} (\textcolor[RGB]{254,76,97}{+3.4})& \textbf{94.7$\pm$1.5} &\textbf{89.9$\pm$1.2} &\textbf{93.6$\pm$0.8}& \textbf{93.0$\pm$1.4} &\textbf{92.8} (\textcolor[RGB]{254,76,97}{+1.8}) & \textbf{93.3} (\textcolor[RGB]{254,76,97}{+2.6})  \\ \specialrule{0.05em}{0pt}{3pt}
\multicolumn{13}{c}{Average Symmetric Surface
Distance (avg.$\pm$ std., $mm$) $\downarrow$}  \\  \specialrule{0.05em}{3pt}{0pt}
UMMKD\!\!\!&\!\!\!\scriptsize[\textcolor[RGB]{52 152 219}{TMI20}] & -&  -& -& -&-& -&- &- &- &-& - \\ 
\multicolumn{2}{c|}{Backbone} & 1.19$\pm$0.91 &   1.18$\pm$0.82&1.84$\pm$1.06 & 1.10$\pm$0.78 & 1.33& 1.20$\pm$0.68& 1.27$\pm$0.79 &1.36$\pm$0.94 &1.37$\pm$0.71 &1.30 & 1.31 \\  
\multicolumn{2}{c|}{Baseline} & 1.12$\pm$0.75&  0.98$\pm$0.68& 1.60$\pm$0.93&  1.05$\pm$0.65& 1.19& 1.07$\pm$0.52& 1.19$\pm$0.76 &1.22$\pm$0.80 &1.23$\pm$0.64 &1.18 & 1.18 \\ \hline
\multicolumn{2}{c|}{Joint Training} & 1.03$\pm$0.62&  0.85$\pm$0.51& 1.87$\pm$0.84 & 0.96$\pm$0.58& 1.18 (\textcolor[RGB]{254,76,97}{-0.01})& 1.19$\pm$0.56 & 1.32$\pm$0.73 &1.27$\pm$0.85 &1.34$\pm$0.67 &1.28 (\textcolor[RGB]{0 205 102}{+0.10}) & 1.23 (\textcolor[RGB]{0 205 102}{+0.05}) \\ 
\multicolumn{2}{c|}{Ours(w/o CR)} & 0.94$\pm$0.58 &  0.75$\pm$0.37 & 1.37$\pm$0.61 & 0.82$\pm$0.43 & 0.97 (\textcolor[RGB]{254,76,97}{-0.22})& 1.01$\pm$0.49& 1.18$\pm$0.64 &1.03$\pm$0.69 &1.15$\pm$0.53 &1.09 (\textcolor[RGB]{254,76,97}{-0.09}) & 1.03 (\textcolor[RGB]{254,76,97}{-0.15}) \\ 
\multicolumn{2}{c|}{\textbf {Ours(w CR)}} & \textbf{0.87$\pm$0.29} &  \textbf {0.58$\pm$0.17} & \textbf {0.84$\pm$0.32} & \textbf{0.72$\pm$0.24} & \textbf {0.75} (\textcolor[RGB]{254,76,97}{-0.44}) & \textbf {0.83$\pm$0.36}& \textbf {0.56$\pm$0.23} & \textbf {0.85$\pm$0.39} &\textbf {0.83$\pm$0.37} & \textbf {0.77} (\textcolor[RGB]{254,76,97}{-0.41}) & \textbf {0.76} (\textcolor[RGB]{254,76,97}{-0.42})  \\
\hline
 \end{tabular}}
  \caption{The results of abdominal multi-organ segmentation by using 3D Transformer.}
 \label{tbl:organ}
 \vspace{-0.7cm}
\end{table*}

\subsection{Abdominal Multi-organ Segmentation}
\textbf{3D Configuration.}
We adopt public CT data from \cite{landman20152015} with $30$ patients, and the T2-SPIR MRI data from the ISBI 2019 CHAOS Challenge \cite{kavur2021chaos}.
%
% Since the UMMKD method does not report the detailed partitioning of these two datasets, we attempt to align their practices for a more fair comparison.
% %
% The UMMKD method \cite{dou2020unpaired}, in particular, employs only $9$ MRI scans from the MRI dataset and removes a low-quality CT scans from the CT dataset.
% As a result, we randomly select $9$ MRI cases from all $20$ MRI scans five times and use the entire CT images from CT dataset to demonstrate the robustness of our method, and then average the results of the five randomized experiments to compare with their method. In accordance with UMMKD, we randomly divided each data modality into $70\%$ for training, $10\%$ for validation, and $20\%$ for testing.
% %
% Following UMMKD's pre-processing methods, we first resample them into around $1.5 \times 1.5 \times 8.0~\rm{mm}^3$ with a size of $256 \times 256$ in the coronal plane to eliminate the huge difference in voxel-spacing between two datasets, and then perform intensity normalization to zero mean and unit variance for each modality.
We employ a 3D U-shaped Transformer named nnFormer \cite{zhou2021nnformer} as the backbone. The detailed dataset and network descriptions are in \textbf{Appendix}~\ref{sec:imple}

\textbf{Main Results.} 
% Here we implement the transformer-based 3D nnFormer as our \text{Backbone}. 
%
Here, the \text{Baseline} model is the \text{Backbone} with $\mathcal{L}_{\rm{aux}}$, which is the same as the procedure in the 2D case.
The Dice values raise to $90.3\%$ on CT and $91.0\%$ on MRI, which is shown in Table~\ref{tbl:organ}.
Likewise, sharing the entire Backbone model in Joint Training causes a slight segmentation performance drop, i.e. Dice of 90.6\% on CT and 90.5\% on MRI. By integrating modality-specific activation, \text{Ours (w/o CR)} improves the Dice values to $91.4\%$ on CT and $91.6 \%$ on MRI, outperforming \text{Backbone}, \text{Baseline}, and \text{Joint Training} by a large margin.
Furthermore, by using the $\mathcal{L}_{\rm{mcr}}$ and $\mathcal{L}_{\rm{icr}}$, our full multi-modality learning scheme marked as \text{Ours (w/ CR)} achieves the best segmentation results of $93.3\%$ overall mean Dice and $0.76 \rm{mm}$ overall mean ASD. 
Compared with the counterpart, our scheme also outperform UMMKD by a significant margin, \textit{i.e.} the Dice value of $1.3\%$ on CT and $3.5\%$ on MRI. 
The visual segmentation results for quantitative comparison are in \textbf{Appendix}~\ref{fig:cardiac_abdominal}.

% \subsection{Qualitative Results.} 
% The qualitative comparisons contain the visual segmentation results of different models for two multi-modality tasks, as shown in Fig.~\ref{fig:cardiac} (Left for Cardiac; Right for Abdominal).

% We exhibit the visual segmentation results of various approaches for , as shown in Fig.~\ref{fig:cardiac}(Right).
%放一张大表

\noindent
\section{Conclusion}
This paper studies how to train the single segmentation model to conduct unpaired multi-modal medical image predictions.
A novel plug-and-play External Attention Module (EAM) is introduced to regulate the backbone network to obey the structured semantic consistency for different modalities, \textit{i.e.} modality-specific class representations and image-specific inter-class correlations.
In the test phase, the EAMs can be removed, maintaining the simplicity of the network.
%superior 
%
Extensive experimental results show the effectiveness of our method.
%
%
%especially by using limited training samples for the specific modality. 
%
%illustrating the capacity of the proposed framework on modal mutual promotion to reduce modality discrepancy.
%

\midlacknowledgments{The work is supported in part by National Key R\&D Program of China under grant No. 2022ZD0116004, by the Young Scientists Fund of the National Natural Science Foundation of China under grant No. 62106154, by Natural Science Foundation of Guangdong Province, China (General Program) under grant No.2022A1515011524, by Guangdong Basic and Applied Basic Research Foundation under Grant No. 2017A030312006, by Shenzhen Science and Technology Program ZDSYS20211021111415025, and by the Guangdong Provincial Key Laboratory of Big Data Computing, The Chinese University of Hong Kong (Shenzhen). }
\bibliography{midl23_181}

%%%%%%%%%%%%%%%%%%%%%%%%%%%%%%
%%%%%%%%%%%%%%%%%%%%%%%%%%%
%%%%%%%%%%%%%%%%%%%%%%%%%%

\appendix

\section{Related Work}

\subsection{Domain adaptation.} In medical image analysis, severe domain shift has been a long-standing obstacle to knowledge transfer between unpaired modalities obtained by different physical principles of imaging. To address this problem, research works on domain adaptation of models either in a semi-supervised \cite{zhu2021semi} or unsupervised \cite{chen2020unsupervised,huo2018synseg} manner, aiming to effectively improve cross-modality representation learning. 
To use data more effectively, some works also focus on full-supervised multi-modality learning, transferring the knowledge that can promote each other between different modalities. 
In \cite{valindria2018multi}, Valindria \textit{et al.} discussed the effectiveness of various dual-stream architectures, demonstrating that the domain shift between two unpaired modalities limits mutual information sharing.
Following this, Dou \textit{et al.} \cite{dou2020unpaired} proposed a multi-modal learning approach by employing modality-specific
internal normalization layers to compute respective statistics of each modality, and conduct cross-modality knowledge distilling to reduce the gap of prediction distributions between modalities.
Instead of carefully constructing multi-modal networks with different feature fusion strategies, our work aims to design the external plug-and-play module to help a single model to establish the structured semantic consistency of different modalities, realizing multi-modal predictions.

\subsection{Vision Transformer.} Currently, the studies about vision Transformer \cite{dosovitskiy2020image} also achieve great progress in image analysis and understanding. ViT is a convolution-free network architecture, which directly employs the attention mechanism to capture the long-range dependence of a sequence of non-overlapping image patches, 
Followed by ViT, Touvron \textit{et al.} \cite{touvron2021training} proposed DeiT that introduced a distillation strategy for Transformer to help with ViT training. And many other ViT variants are also proposed \cite{chen2021crossvit,wang2021pyramid,rao2021dynamicvit,zheng2021rethinking,liu2021swin} which achieve promising performance compared with its counterpart CNNs on various vision tasks, such as image classification and semantic segmentation. 
Inspired by SwinTransformer, Cao \textit{et al.} \cite{cao2021swin} proposed a pure U-shaped transformer named Swin-Unet, where the architecture utilizes SwinTransformer block as the basic unit to improve its capacity of feature representation for 2D medical image segmentation. Furthermore, to explore Transformer's ability to learn volumetric representations from 3D medical volumes, Zhou \textit{et al.} \cite{zhou2021nnformer} proposed nnFormer to interleave convolution and self-attention for medical volumetric segmentation, achieving tremendous progress over previous transformer-based medical segmentation methods.

\section{Methodology Details}
\subsection{Transformer Block}
\label{sec:transformer}
%
%Following Vision Transformer \cite{vaswani2017attention}, the Transformer Block includes the basic multi-head attention, layer normalization, and feed forward layer. 
%
%
Motivated by the LayerScale \cite{touvron2021going}, 
we introduce modality-aware channel-wise multiplication on the output of each residual operation in the Transformer block, with the goal of calibrating modality-specific activation in the channel dimension to narrow the discrepancy between representations from different modalities.
Given the modality-specific class embeddings $\mathbf{Q} \in  \mathbb{R}^{Z \times 4C}$ of a certain modality, where $Z$ is the total number of classes and is constant across all modalities, and $4C$ denotes the channel dimension, we aggregate its semantic information by using linear projection to produce modality-specific channel weight $\mathbf{\Omega} \in  \mathbb{R}^{4C}$.
We then project $\mathbf{\Omega}$ to the corresponding feature dimensions (\textit{e.g.,} from $4C$ to $D$ for the specific scale),
and the $\rm{diag}(\cdot)$ operation is adopted to generate diagonal matrix to calibrate modality-specific activation as,
%
%In our single Transformer, the modality-specific weight $\mathbf{\Omega}$ will act as a diagonal matrix to interact with multi-scale patch tokens by linearly projecting to corresponding sub-spaces:
%
\begin{gather}
    \mathbf{\Omega}=\mathbf{w}_1\mathbf{Q},~
    \mathbf{\Phi}_1= {\rm{diag}}( \mathbf{\Omega}\mathbf{W}_2 ),~
    \mathbf{\Phi}_2= {\rm{diag}}( \mathbf{\Omega}\mathbf{W}_3 ),\\
    \mathbf{X}_{l}^{\prime}=\mathbf{X}_{l}+\mathbf{\Phi}_1 \circledast {\rm{MSA}}(~{\rm{Norm}}(\mathbf{X}_{l})~),\\
    \mathbf{X}_{l+1}=\mathbf{X}_{l}^{\prime}+\mathbf{\Phi}_2 \circledast {\rm{FFN}}(~{\rm{Norm}}(\mathbf{X}_{l}^{\prime})~),
\end{gather}
where $\mathbf{w}_1 \in  \mathbb{R}^{Z}$, $\mathbf{W}_2 \in  \mathbb{R}^{4C \times D}$, $\mathbf{W}_3 \in  \mathbb{R}^{4C \times D}$, 
%
%$\mathbf{\Psi}_1$ is ${\rm{diag}}(\varphi_{l,1},...,\varphi_{l,D})$ and $\mathbf{\Psi}_2$ is ${\rm{diag}}(\varphi_{l,1}^{\prime},...,\varphi_{l,D}^{\prime})$. 
%
$\rm MSA(\cdot)$ and $\rm FFN(\cdot)$ denote the multi-head attention layer and the feed-forward layer respectively. $\rm Norm(\cdot)$ indicates the LayerNorm operation, and $\circledast$ is the channel-wise multiplication.
$\mathbf{X}_{l}$ and $\mathbf{X}_{l+1}$  denote the input and output of $(l+1)$-th transformer block. 
As shown in Fig.~\ref{fig:frmework}, we feed the modality-specific class embeddings into the Transformer decoder,
%
%In practice, we find that the modality-specific channel-wise multiplication can also calibrate the universal representations extracted in the Transformer encoder. 
%
and we implement the above scheme in all Transformer blocks to improve accuracy even further. 
\textbf{It is worth noting} that such modality-aware channel-wise recalibration is an optional operation if we require to drop the modality-aware query as well in inference. 
%
% Thus, we introduce the above scheme into all the Transformer blocks to further improve the accuracy.
% %
% It is also worth noting that such modality-aware channel-wise recalibration is optional. 
% %
% If we require to drop the modality-aware query as well in inference, 
% we could omit such scheme and directly employ original Transformer blocks proposed in~\cite{cao2021swin, zhou2021nnformer}.

% It should also be noted that such modality-specific channel-wise recalibration is optional. If we need to remove the modality-specific class embeddings during inference, we could skip this scheme and instead use the original Transformer blocks proposed in \cite{cao2021swin,zhou2021nnformer}.

\subsection{External Attention Module}
\begin{figure*}[t]
\centering
		\includegraphics[width=1.0\textwidth]{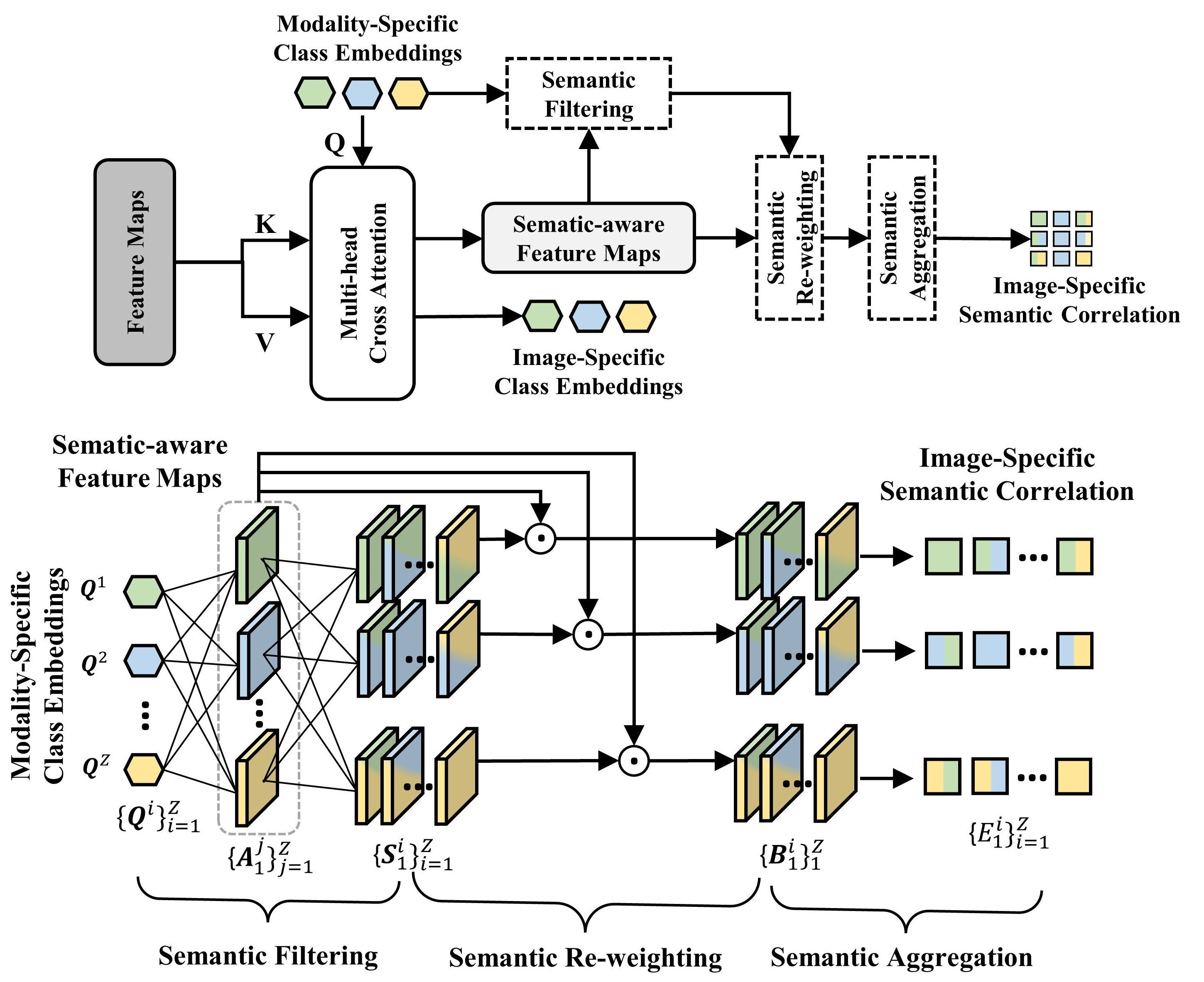}
	% \vspace{-10pt}	
 \caption{
	%
	%which adopts modality-aware query as the input to realize the calibration of modality-specific activation,  
	%
	 External Attention Module (EAM) 
 %  for the Stage-1 of Transformer Decoder, which adopts \textit{modality-specific class embeddings} and feature maps at scale-1 as the input. 
	% %
	% In practice, we replace \textit{modality-specific class embeddings} with \textit{image-specific class embeddings} for Stage-2 and Stage-3, which is illustrated in Fig.~\ref{fig:frmework}. 
	%
	} 
	\label{fig:EAM}
% \vspace{-0.5cm}
\end{figure*}

In the main article, we design the External Attention Module (EAM) to update the learnable class embeddings from modality-specific to image-specific, and extract the semantic correlations of a specific image at multiple scales. Fig.~\ref{fig:EAM} shows the details of EAM for the Stage-1 of Transformer Decoder, which adopts \textit{modality-specific class embeddings} and feature maps at scale-1 as the input. In practice, we replace \textit{modality-specific class embeddings} with \textit{image-specific class embeddings} for Stage-2 and Stage-3, which is illustrated in Fig.~\ref{fig:frmework}. 

% \section{Experiments}
\section{Implementation Details}
\label{sec:imple}
\subsection{Datasets}
\textbf{Cardiac Substructure Segmentation.} 
% we take the same partitioning (\textit{i.e.} training, validation and test) and pre-processing of the Multi-Modality Whole Heart Segmentation dataset \cite{zhuang2019evaluation}  as the UMMKD reported for a fair comparison. Specifically, we align with them to split each modality into $70\%$ for training, $10\%$ for validation, and $20\%$ for testing. 
We employ the Multi-Modality Whole Heart Segmentation Challenge 2017 dataset \cite{zhuang2019evaluation} to perform  multi-class cardiac structure segmentation. The dataset is composed of unpaired $20$ CT and $20$ MRI scans collected from various patients and sites.
We intend to train the segmentation network to recognize four organs for each modality: \texttt{left ventricle myocardium} (LVM), \texttt{left atrium blood cavity} (LAC), \texttt{left ventricle blood cavity} (LVC), and \texttt{ascending aorta} (AA). 
Since the UMMKD approach \cite{dou2020unpaired} achieves state-of-the-art performance, we take the same partitioning (\textit{i.e.} training, validation, and test) and pre-processing of the dataset as the UMMKD reported for a fair comparison.
Specifically, we align with them to split each modality into $70\%$ for training, $10\%$ for validation, and $20\%$ for testing. Before feeding into the network, we first resample both modalities to isotropic 1$\rm{mm}^3$ and crop their central heart region by a 3D bounding box which has a fixed coronal plane size of $256 \times 256$. 
For each 3D cropped image, we remove the top $2\% $ of its intensity histogram to reduce artifacts, which is then normalized to zero mean and unit variance.

\textbf{Abdominal Multi-organ Segmentation.} We adopt public CT data from \cite{landman20152015} with $30$ patients, and the T2-SPIR MRI data from the ISBI 2019 CHAOS Challenge \cite{kavur2021chaos} with $20$ volumes. 
We focus on adapting the segmentation network to delineate four abdominal organs: \texttt{liver}, \texttt{right kidney} (R-kdy), \texttt{left kidney} (L-kdy), and \texttt{spleen}.
Since the UMMKD method does not report the detailed partitioning of these two datasets, we attempt to align their practices for a more fair comparison.
The UMMKD method \cite{dou2020unpaired}, in particular, employs only $9$ MRI scans from the MRI dataset and removes a low-quality CT scan from the CT dataset.
As a result, we randomly select $9$ MRI cases from all $20$ MRI scans five times and use the entire CT images from the CT dataset to demonstrate the robustness of our method and then average the results of the five randomized experiments to compare with their method. In accordance with UMMKD, we randomly divided each data modality into $70\%$ for training, $10\%$ for validation, and $20\%$ for testing.
% Similarly, we follow the UMMKD approach \cite{dou2020unpaired} to data partitioning and preprocessing for two abdominal datasets.
%
Following UMMKD's pre-processing methods, we first resample them into around $1.5 \times 1.5 \times 8.0~\rm{mm}^3$ with a size of $256 \times 256$ in the coronal plane to eliminate the huge difference in voxel-spacing between two datasets, and then perform intensity normalization to zero mean and unit variance for each modality. 
\subsection{Evaluation Metrics}
We assess segmentation performance using the Volume Dice Coefficient (Dice, $\%$) and Average Symmetric Surface
Distance (ASD, $mm$) metrics, calculating the average and standard deviation of segmentation results for each class \cite{dou2020unpaired}. Note that the final segmentation results (mean and standard variance) in the abdominal multi-organ segmentation task are calculated from the results of all the test samples in the five random tests.

\subsection{Model Configuration}
As in the main article, we implement 2D and 3D Transformers with various network architectures as the backbone to demonstrate the flexibility and general efficacy of our method. Specifically, we adopt a 2D Transformer-based U-shaped Encoder-Decoder architecture named Swin-Unet \cite{cao2021swin} for the multi-class cardiac structure segmentation. For the abdominal multi-organ segmentation, we employ a 3D U-shaped Transformer with a volume-based self-attention mechanism and strided convolution named nnFormer \cite{zhou2021nnformer}. We report the detailed network architecture in Table.~\ref{tbl:config}.
\begin{table*}[!t]

\centering
\resizebox{\textwidth}{!}{
\begin{tabular}{ccc} 
\toprule [1.5 pt]
 \multicolumn{1}{c}{Layer} & \multicolumn{1}{c}{2D Transformer \cite{cao2021swin}} & \multicolumn{1}{c}{3D Transformer \cite{zhou2021nnformer}}   \\ \hline 
 Transformer Block & Swin Transformer Block \cite{liu2021swin} & Volume-based Self-attention\\
Down-sampling & Patch Merging  &  Strided Convolution\\
Up-sampling & Patch Expanding  &  Strided Convolution\\
$[G_1, G_2, G_3, G_4]$ & [2, 2, 2, 2] & [2, 2, 2, 2]\\
$[L_1, L_2, L_3, L_4]$ & [2, 2, 2, 2] & [2, 2, 2, 2]\\
Patch Size & $4\times4$& $4\times4\times2$\\
\hline
 \end{tabular}}
  \caption{\label{tbl:config}Detailed configurations of 2D Transformer and 3D Transformer architectures corresponding to Fig.~\ref{fig:frmework}, where $\{G_{i}\}_{i=1}^4$ and $\{L_{i}\}_{i=1}^4$ denote the basic number of Transformer blocks for the Encoder and Decoder respectively.}
% \vspace{-0.5cm}
\end{table*}

\begin{figure*}[htbp]
	\centering
		\includegraphics[width=0.3\linewidth]{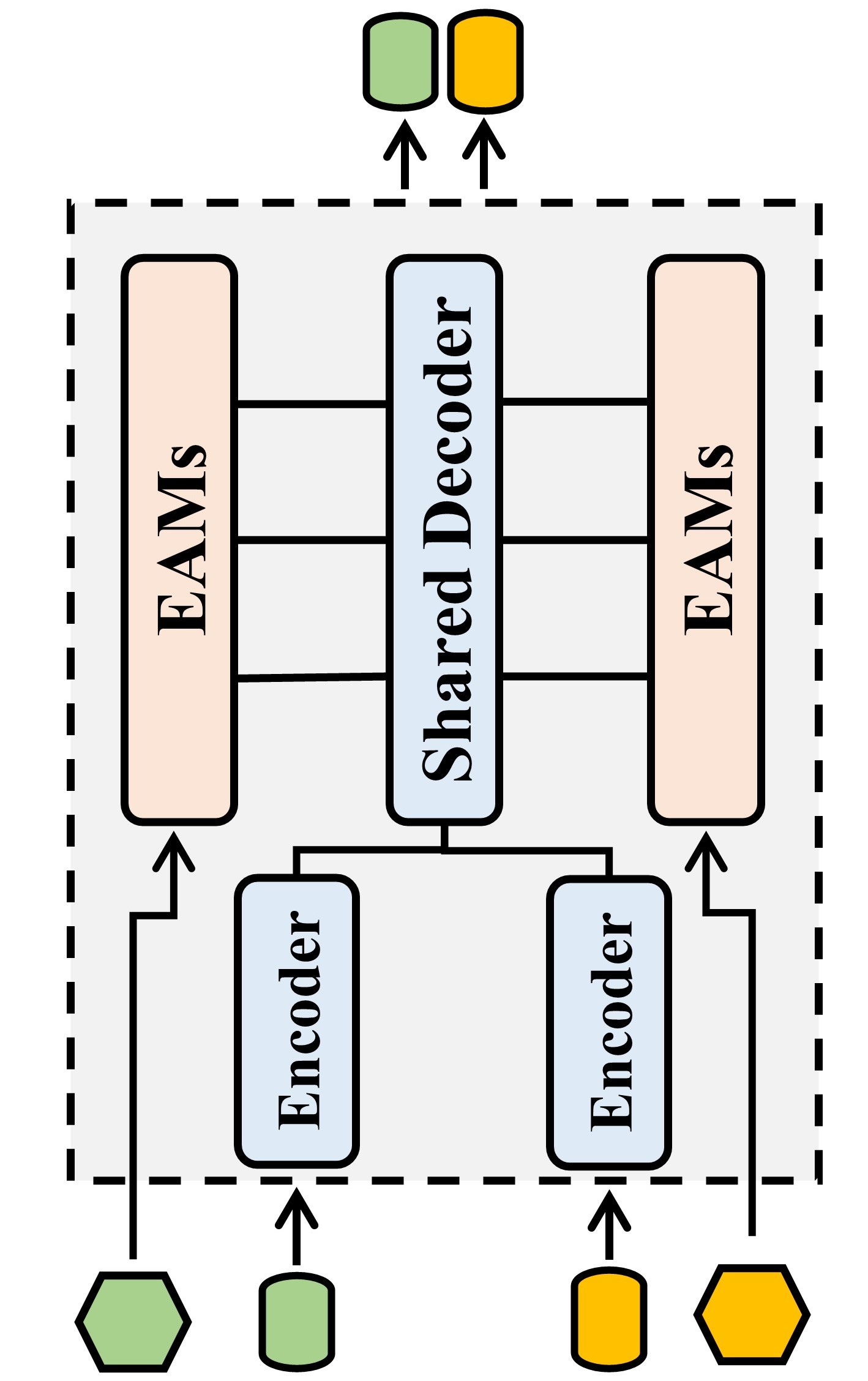}
  \hfill
		\includegraphics[width=0.3\linewidth]{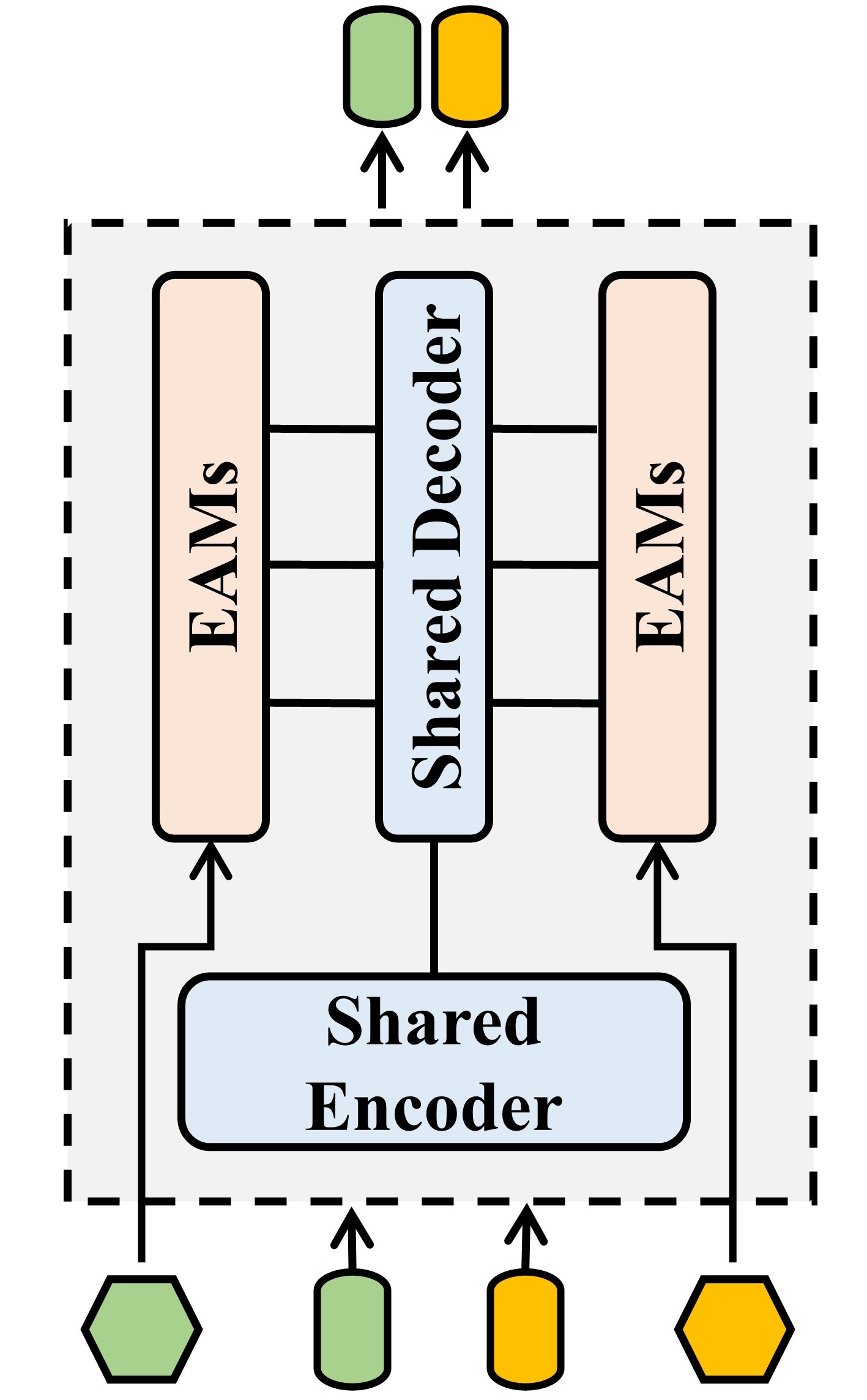}
  \hfill
		\includegraphics[width=0.3\linewidth]{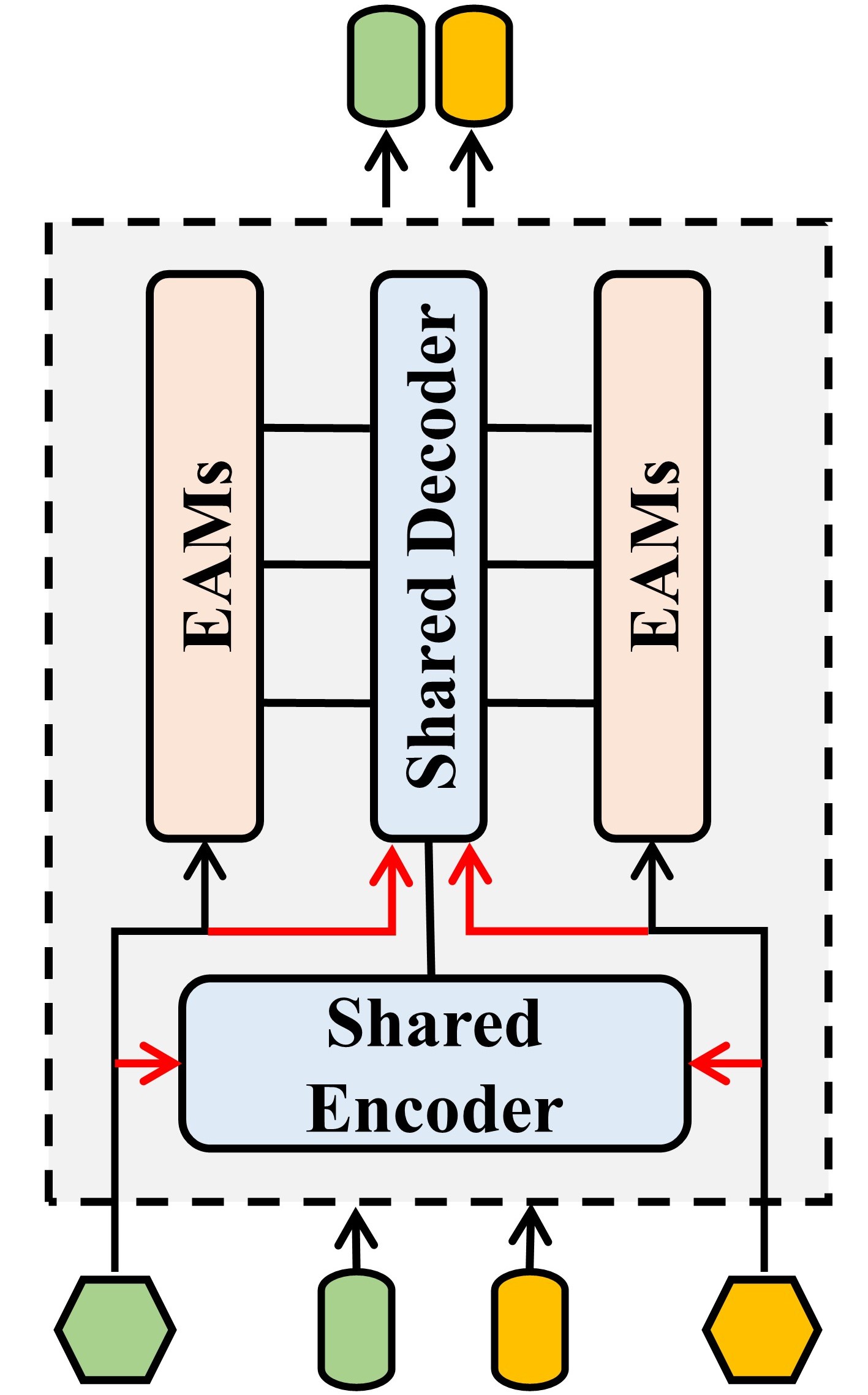}
	\caption{Three versions of Joint architecture: (a) Using modality-specific Encoder and EAMs while sharing Decoder; (b) Sharing both Encoder and Decoder while keeping EAMs as modality-specific; (c) Incorporating the modality-aware channel-wise multiplication mechanism into each transformer block of shared Encoder and Decoder in the architecture (b).
	Note that the input hexagons and rectangles denote modality-specific class embeddings and patch tokens, respectively.
	}
	\label{fig:joint}
\end{figure*}

% 2D and 3D backbone table
% \section{Supplementary Experiment}
\section{Few-shot Domain Adaptation}
\subsection{Experiment Setting} 
To assess the effectiveness of our method when training with limited samples of one modality, we randomly select $1$ or $3$ samples from one modality's (CT or MRI) training set while training the model with all training data from another modality.
We conduct all of the few-shot domain adaptation experiments on MICCAI-MMWHS dataset. 

\subsection{Main Results} 
As shown in Table~\ref{tbl:few-shot}, when our multi-modal learning framework is trained with only $1$ annotated MRI image and all CT images, the value of mean Dice for MRI is only slightly lower than {Baseline} (\textit{i.e.} $0.4\%$), which is trained with all MRI images alone.
This demonstrates that when using complete CT training data, the proposed method can eliminate the impact of insufficient MRI training samples.
In such a case, the mean Dice of CT improves to $92.7\%$ and the mean ASD decreases to $1.41\rm{mm}$, 
evidencing that the feature representation of CT could also be enhanced by using only $1$ MRI image.
Moreover, when the number of MRI training samples increases to $3$, the mean Dice values improve to $93.3\%$ on CT and $86.9\%$ on MRI, surpassing the {Baseline} model.

When training with only $1$ CT image and all of MRI images using our scheme, the mean Dice of CT drops by $0.7\%$ compared with {Baseline} model which is only trained with all of CT images. 
Also, giving $1$ annotated CT image improves the mean Dice of MRI by $1.1\%$ and decreases the mean ASD of MRI by $0.17\rm{mm}$, highlighting the mutual promotion between CT and MRI. %
Further, using $3$ CT images raises the mean Dice to $91.5\%$ on CT and $88.0\%$ on MRI, defeating UMMKD by $2.0\%$ on MRI while maintaining competitive performance on CT.

\begin{table*}
\centering
 \resizebox{\textwidth}{!}{
\begin{tabular}{l|ccccl|ccccl|l} 
\toprule [1.5 pt]
\multirow{2}{*}{Methods}& \multicolumn{5}{c|}{Cardiac CT} &
\multicolumn{5}{c|}{Cardiac MRI} & \multirow{2}{*}{Overall Mean} \\
\cline{2-11} 
   &  LVM & LAC & LVC & AA & Mean &  LVM & LAC & LVC & AA & Mean & \\ \specialrule{0.05em}{0pt}{3pt}
\multicolumn{12}{c}{Dice Coefficient (avg.$\pm$ std., $\%$) $\uparrow$}  \\ \specialrule{0.05em}{3pt}{0pt}
% UMMKD  & 88.5&  91.5& 93.1& 93.6& 91.7& 80.8& 86.5 &93.6 &83.1 &86.0 &88.8 \\
% Ours & 90.9 &  94.8& 94.5& 95.9& 94.0& 81.6& 89.6 &94.4 &89.2 &88.7 & 91.4\\ 
Baseline & 90.6$\pm$2.8&  92.6$\pm$2.8& 93.2$\pm$2.5& 88.9$\pm$3.4& 91.3 & 80.9$\pm$4.0& 86.3$\pm$3.8 &92.9$\pm$2.3 &85.8$\pm$3.5 &86.5 & 88.9 \\
Ours (w/ CR) &90.9$\pm$2.0&  94.8$\pm$1.6& 94.5$\pm$2.1& 95.9$\pm$1.4& 94.0 & 81.6$\pm$2.5& 89.6$\pm$3.3 &94.4$\pm$1.3 &89.2$\pm$2.8 &88.7 &91.4\\
\hline 
One annotated MRI & 90.7$\pm$2.7 &  93.1$\pm$2.5 & 93.3$\pm$2.8& 93.7$\pm$2.4 & 92.7 (\textcolor[RGB]{254,76,97}{+1.4}) & 78.8$\pm$5.8& 85.2$\pm$4.6 &92.5$\pm$3.2 &87.9$\pm$4.1 &86.1 (\textcolor[RGB]{0 205 102}{-0.4}) & 89.4 (\textcolor[RGB]{254,76,97}{+0.5})\\
Three annotated MRI & 90.4$\pm$2.3 &  93.7$\pm$2.0 & 93.9$\pm$2.7 & 95.0$\pm$1.9 & 93.3 (\textcolor[RGB]{254,76,97}{+1.2}) & 80.6$\pm$4.7 & 86.0$\pm$4.1 &92.6$\pm$2.9 &88.2$\pm$3.3 &86.9 (\textcolor[RGB]{254,76,97}{+0.4}) &90.1 (\textcolor[RGB]{254,76,97}{+1.4}) \\\hline 
One annotated CT & 88.2$\pm$3.4 & 91.2$\pm$3.5 & 91.4$\pm$2.2 & 91.6$\pm$3.3 & 90.6 (\textcolor[RGB]{0 205 102}{-0.7}) & 80.8$\pm$4.2& 87.9$\pm$4.3 &93.1$\pm$2.1 &88.6$\pm$2.9 &87.6 (\textcolor[RGB]{254,76,97}{+1.1}) &89.1 (\textcolor[RGB]{254,76,97}{+0.2}) \\
Three annotated CT & 88.9$\pm$3.1 & 91.7$\pm$2.9 & 92.4$\pm$2.3 & 92.8$\pm$2.8 & 91.5 (\textcolor[RGB]{254,76,97}{+0.2}) & 81.1$\pm$3.7 & 88.4$\pm$3.5  &93.5$\pm$1.8 &88.9$\pm$2.7 &88.0 (\textcolor[RGB]{0 205 102}{+1.5}) &89.7 (\textcolor[RGB]{254,76,97}{+0.8})\\
\specialrule{0.05em}{0pt}{3pt}
\multicolumn{12}{c}{Average Symmetric Surface
Distance (avg.$\pm$ std., $mm$) $\downarrow$}  \\ \specialrule{0.05em}{3pt}{0pt} 
% UMMKD & -&  -& -& -& -& -& - &- &- &- &-\\
% Ours & 1.31&  1.49&1.22& 1.00& 1.26& 1.55& 1.24 &1.27 &2.01 &1.52 & 1.39  \\
Baseline & 1.49$\pm$0.33&  1.84$\pm$0.44& 1.38$\pm$0.35& 1.46$\pm$0.28& 1.54& 1.71$\pm$1.43& 1.37$\pm$0.64 &1.46$\pm$0.89 &2.69$\pm$1.27 &1.86 & 1.70 \\ 
Ours (w/ CR)&1.31$\pm$0.27&  1.49$\pm$0.38& 1.22$\pm$0.27& 1.00$\pm$0.24& 1.26& 1.55$\pm$0.78& 1.24$\pm$0.34 &1.27$\pm$0.32 &2.01$\pm$0.95 &1.52 & 1.39  \\
\hline
One annotated MRI & 1.34$\pm$0.34 &  1.68$\pm$0.48 & 1.37$\pm$0.36 & 1.25$\pm$0.30 & 1.41 (\textcolor[RGB]{254,76,97}{-0.13})& 1.89$\pm$1.59 & 2.04$\pm$0.93 &1.63$\pm$1.01 &2.36$\pm$1.18 &1.98 (\textcolor[RGB]{0 205 102}{+0.12}) & 1.70 (+0.00)\\
Three annotated MRI &  1.41$\pm$0.37 & 1.55$\pm$0.41  & 1.34$\pm$0.47  & 1.16$\pm$0.26 & 1.37 (\textcolor[RGB]{254,76,97}{-0.17}) & 1.73$\pm$1.28 & 1.82$\pm$0.87 & 1.51$\pm$0.83 & 2.24$\pm$1.06& 1.83 (\textcolor[RGB]{254,76,97}{-0.04})& 1.60 (\textcolor[RGB]{254,76,97}{-1.00}) \\\hline 
One annotated CT & 1.86$\pm$0.58 & 2.13$\pm$0.86 & 1.92$\pm$0.92 & 1.31$\pm$0.35 & 1.81 (\textcolor[RGB]{0 205 102}{+0.27}) & 1.75$\pm$1.31 & 1.44$\pm$0.47 & 1.42$\pm$0.75 & 2.08$\pm$1.02 & 1.67 (\textcolor[RGB]{254,76,97}{-0.19}) & 1.74 (\textcolor[RGB]{0 205 102}{+0.04}) \\
Three annotated CT & 1.69$\pm$0.44 & 1.95$\pm$0.71 & 1.61$\pm$0.59 & 1.26$\pm$0.37 & 1.63 (\textcolor[RGB]{0 205 102}{+0.09}) & 1.62$\pm$0.94  & 1.34$\pm$0.52 & 1.36$\pm$0.49 & 2.14$\pm$1.12 & 1.62 (\textcolor[RGB]{254,76,97}{-0.24}) & 1.62 (\textcolor[RGB]{254,76,97}{-0.08}) \\
 \hline
 \end{tabular}}
\caption{The results of \textbf{few-shot domain adaptation} on cardiac segmentation by using 2D network.}
 \label{tbl:few-shot}
\end{table*}

\begin{figure*}[ht]
	\begin{center}
		\includegraphics[width=1\linewidth]{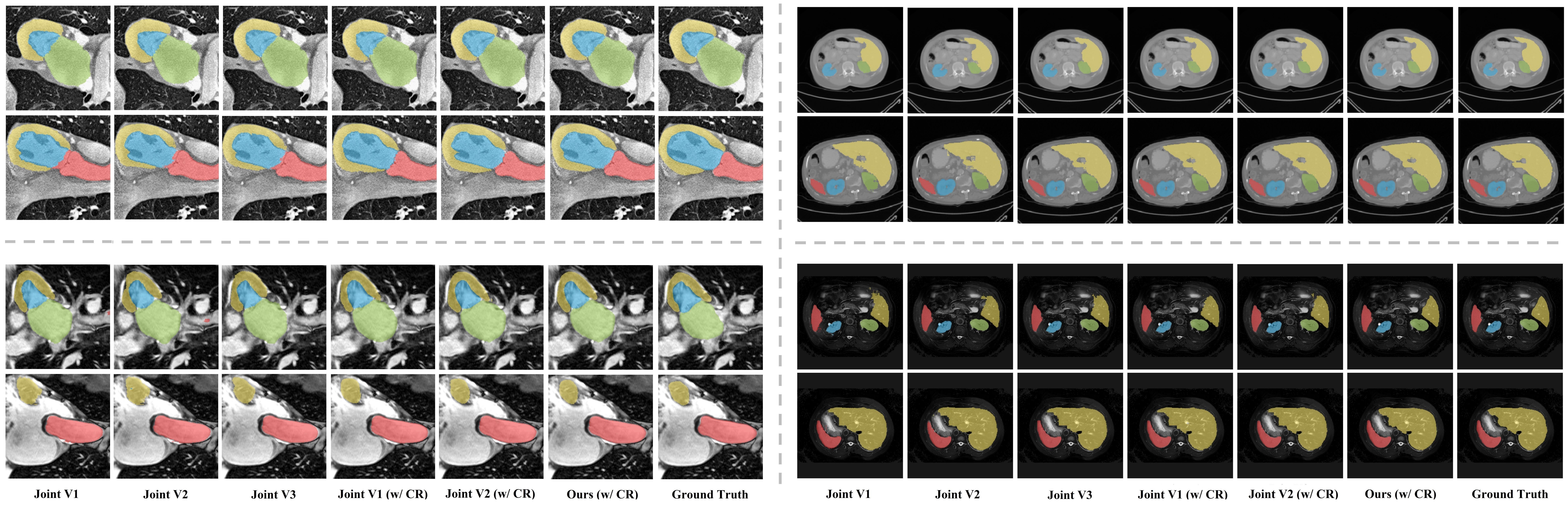}
	\end{center}
	% \vspace{-10pt}
	\caption{Visualization results on cardiac segmentation task and abdominal multi-organ task. \textbf{Left:} AA, LAC, LVC and LVM. \textbf{Right:} Spleen, R-kdy, L-kdy and Liver. The corresponding colormaps are \textbf{\textcolor[RGB]{238,37,36}{red}}, \textbf{\textcolor[RGB]{145, 193,  62}{green}}, \textbf{\textcolor[RGB]{29, 162, 220}{blue}} and \textbf{\textcolor[RGB]{254, 232,  81}{yellow}}, respectively.} 
\label{fig:cardiac_abdominal}
% \vspace{0.1cm}
\end{figure*}

\section{Ablation Studies}

\subsection{Three Versions of Joint Architecture}
To thoroughly evaluate the effectiveness of the different joint architectures and two consistency regularization terms of our method, we provide three degenerate models of the proposed method with or without the two consistency regularization losses. The detailed architectures are shown in Fig.~\ref{fig:joint}. We report the six experimental settings that conduct the fixed 2D or 3D Transformer and hyper-parameters: (1) \textbf{Joint V1} that is a joint Transformer with modality-specific encoder and EAMs, shared decoder, which is a conventional multi-modal learning architecture \cite{nie2016fully} as shown in Fig.~\ref{fig:joint}-1; 
2) \textbf{Joint V2} (denoted as \textbf{Joint Training} in the main article) is a joint Transformer with shared encoder and decoder, and modality-specific EAMs, as shown in Fig.~\ref{fig:joint}-2;
3) \textbf{Joint V3} (denoted as \textbf{Ours (w/o CR)} in the main article) is trained based on Joint V2, implementing the modality-aware channel-wise multiplication mechanism in each transformer block of the shared encoder and decoder, as shown in Fig.~\ref{fig:joint}-3;
4) \textbf {Joint V1 (w/ CR)} is trained based on {Joint V1} Transformer with two types of consistency terms $\mathcal{L}_{\rm{mcr}}$, $\mathcal{L}_{\rm{icr}}$; 
5) \textbf {Joint V2 (w/ CR)} is trained based on {Joint V2} Transformer with two types of consistency terms $\mathcal{L}_{\rm{mcr}}$, $\mathcal{L}_{\rm{icr}}$; 
6) \textbf{Ours (w/ CR)} that is our full cross-modal learning strategy by adding two types of consistency terms $\mathcal{L}_{\rm{mcr}}$, $\mathcal{L}_{\rm{icr}}$.

\subsubsection{Cardiac Substructure Segmentation.}
As shown in Fig.~\ref{fig:joint}-1, based on the {Baseline} model, {Joint V1} exemplifies a conventional multi-modal Transformer architecture by using modality-specific encoders and EAMs while sharing decoders.
This model raises the Dice value to $91.9\%$ for CT and $86.8\%$ for MRI. 
This clearly illustrates the Transformer's ability to deal with multi-modal data with large appearance differences at the same time, as well as the potential for mutual promotion between different modalities by sharing some modules.
%
%not only the sufficient capacity of the Transformer to analyze two modalities with large differences in data distribution, but also the potentiality of reasonably shared parameters for feature fusion to mutually promote each modality. 
%
In Fig.~\ref{fig:joint}-2, {Joint V2}  targets to further share decoder between different modalities to improve the parameter efficiency of multi-modal Transformer.
However, there is a decrease in segmentation results, \textit{i.e.} Dice of $91.5\%$ on CT and $86.2\%$ on MRI. 
This suggests that in such a situation, modality discrepancy has a significant impact on learned feature representations without introducing any semantic consistency constraints.
%
% In Fig.~\ref{fig:joint}-(c), compared with \textit{Joint V2}, we further introduce modality-aware channel-wise multiplication mechanism into each Transformer block of shared encoder and decoder.
% %
% Such scheme improves the segmentation result to $92.8\%$ on CT and $87.4\%$ on MRI, demonstrating the effectiveness of modality-specific activation calibrating.
In Fig.~\ref{fig:joint}-3, we introduce a modality-aware channel-wise multiplication mechanism into each Transformer block of shared encoder and decoder. This scheme improves segmentation results to $92.8\%$ on CT and $87.4\%$ on MRI, demonstrating the efficiency of modality-specific activation calibration.

By leveraging proposed two types of consistency regularization terms, \textit{i.e.} $\mathcal{L}_{\rm{mcr}}$ and $\mathcal{L}_{\rm{icr}}$, the three multi-modal Transformer architectures {Joint V1}, {Joint V2} and {Joint V3} are all boosted. Specifically, The {Joint V1 (w/ CR)} and {Joint V2 (w/ CR)} improve the Dice value by $1.9\%$ and $1.7\%$ on CT respectively. The {Joint V3} with $\mathcal{L}_{\rm{mcr}}$ and $\mathcal{L}_{\rm{icr}}$ is marked as \textbf{Ours (w/ CR)} in Table~\ref{tbl:cardiac}, and it achieves an overall Dice of $91.4\%$ (\textit{i.e.} the average of $94.0\%$ on CT and $88.7\%$ on MRI). In Fig.~\ref{fig:cardiac_abdominal} (Left), we also present a visual representation of the segmentation results for quantitative comparison. 
\begin{table*}
\centering
\resizebox{\textwidth}{!}{
\begin{tabular}{l|ccccl|ccccl|l} 
\hline
\toprule [1.5 pt]
\multirow{2}{*}{Methods} & \multicolumn{5}{c|}{Cardiac CT} &
\multicolumn{5}{c|}{Cardiac MRI} & \multirow{2}{*}{Overall Mean} \\
\cline{2-11} 
    &LVM & LAC & LVC & AA & Mean &  LVM & LAC & LVC & AA & Mean & \\ \specialrule{0.05em}{0pt}{3pt}
\multicolumn{12}{c}{Dice Coefficient (avg.$\pm$ std., $\%$) $\uparrow$}  \\ \specialrule{0.05em}{3pt}{0pt}
Backbone & 90.0$\pm$3.2&  92.5$\pm$2.9& 92.6$\pm$3.0 & 87.4$\pm$3.8& 90.6& 79.9$\pm$4.6 & 85.3$\pm$3.9 &92.0$\pm$2.7 &84.9$\pm$2.9 &85.5& 88.1 \\ 
Baseline & 90.6$\pm$2.8&  92.6$\pm$2.8& 93.2$\pm$2.5& 88.9$\pm$3.4& 91.3 & 80.9$\pm$4.0& 86.3$\pm$3.8 &92.9$\pm$2.3 &85.8$\pm$3.5 &86.5 & 88.9 \\ \hline  
Joint V1 & 89.4$\pm$2.7&  93.3$\pm$3.0& 92.7$\pm$2.9& 92.2$\pm$2.5& 91.9 (\textcolor[RGB]{254,76,97}{+0.6}) &80.5$\pm$4.2& 87.3$\pm$4.3 &92.2$\pm$2.4 &87.0$\pm$3.2 &86.8 (\textcolor[RGB]{254,76,97}{+0.3}) & 89.3 (\textcolor[RGB]{254,76,97}{+0.4}) \\  
Joint V2 & 89.1$\pm$2.8&  93.0$\pm$2.7& 92.8$\pm$3.3& 91.2$\pm$2.6& 91.5 (\textcolor[RGB]{254,76,97}{+0.2}) & 80.2$\pm$3.9& 86.5$\pm$4.5 &92.0$\pm$3.0 &86.1$\pm$3.8 &86.2 (\textcolor[RGB]{0 205 102}{-0.3})  & 88.9 (+0.0) \\ 
Joint V3 & 90.0$\pm$2.3&  93.8$\pm$2.1& 93.4$\pm$2.4& 94.0$\pm$2.0& 92.8 (\textcolor[RGB]{254,76,97}{+1.5}) & 81.0$\pm$3.1& 87.4$\pm$3.6 &93.5$\pm$2.1 &87.8$\pm$3.0 &87.4 (\textcolor[RGB]{254,76,97}{+0.9})  & 90.1 (\textcolor[RGB]{254,76,97}{+1.2}) \\ 
Joint V1 (w/ CR) & 90.2$\pm$2.0&  93.7$\pm$1.8& 93.6$\pm$2.2& 95.1$\pm$1.6& 93.2 (\textcolor[RGB]{254,76,97}{+1.9})& \textbf{81.8$\pm$2.7}& 88.4$\pm$4.0 &93.3$\pm$1.5 &88.6$\pm$2.2 &88.0 (\textcolor[RGB]{254,76,97}{+1.5}) & 90.6 (\textcolor[RGB]{254,76,97}{+1.7}) \\ 
Joint V2 (w/ CR) & 90.5$\pm$2.1&  93.2$\pm$2.2& 93.8$\pm$2.7& 94.4$\pm$1.7& 93.0 (\textcolor[RGB]{254,76,97}{+1.7}) & 81.5$\pm$3.4& 88.0$\pm$4.2 &93.4$\pm$1.9 &88.1$\pm$3.4 &87.8 (\textcolor[RGB]{254,76,97}{+1.3}) & 90.4 (\textcolor[RGB]{254,76,97}{+1.5}) \\ 
\textbf{Ours (w/ CR)} & \textbf{90.9$\pm$2.0} &  \textbf{94.8$\pm$1.6} & \textbf{94.5$\pm$2.1} & \textbf{95.9$\pm$1.4} & \textbf{94.0} (\textcolor[RGB]{254,76,97}{+2.7}) & 81.6$\pm$2.5 & \textbf{89.6$\pm$3.3} & \textbf{94.4$\pm$1.3} & \textbf{89.2$\pm$2.8} & \textbf{88.7} (\textcolor[RGB]{254,76,97}{+2.2}) & \textbf{91.4} (\textcolor[RGB]{254,76,97}{+2.5}) \\ \specialrule{0.05em}{0pt}{3pt}
\multicolumn{12}{c}{Average Symmetric Surface
Distance (avg.$\pm$ std., $mm$) $\downarrow$}  \\ \specialrule{0.05em}{3pt}{0pt}
Backbone & 1.67$\pm$0.46&  1.95$\pm$0.54& 1.43$\pm$0.47& 1.51$\pm$0.41& 1.64& 2.12$\pm$1.57 &1.74$\pm$0.85 & 1.41$\pm$0.81 & 3.74$\pm$1.68& 2.25 & 1.95  \\ 
Baseline & 1.49$\pm$0.33&  1.84$\pm$0.44& 1.38$\pm$0.35& 1.46$\pm$0.28& 1.54& 1.71$\pm$1.43& 1.37$\pm$0.64 &1.46$\pm$0.89 &2.69$\pm$1.27 &1.86 & 1.70 \\ \hline
Joint V1 & 1.63$\pm$0.38&  1.64$\pm$0.40& 1.47$\pm$0.32& 1.19$\pm$0.27& 1.48 (\textcolor[RGB]{254,76,97}{-0.06}) & 1.99$\pm$1.07& 1.36$\pm$0.57 &1.51$\pm$0.73 &2.89$\pm$1.33 &1.94 (\textcolor[RGB]{0 205 102}{+0.08}) & 1.71 (\textcolor[RGB]{0 205 102}{+0.01}) \\  
Joint V2 & 1.58$\pm$0.35&  1.70$\pm$0.44& 1.39$\pm$0.35& 1.33$\pm$0.38& 1.50 (\textcolor[RGB]{254,76,97}{-0.04}) & 1.87$\pm$0.92& 1.47$\pm$0.40 &1.42$\pm$0.55 &3.13$\pm$1.41 &1.97 (\textcolor[RGB]{0 205 102}{+0.11}) & 1.74 (\textcolor[RGB]{0 205 102}{+0.04}) \\
Joint V3 & 1.34$\pm$0.31&  1.63$\pm$0.46& 1.32$\pm$0.27& 1.10$\pm$0.29&1.35 (\textcolor[RGB]{254,76,97}{-0.19}) & 1.84$\pm$0.81& 1.22$\pm$0.53 &1.39$\pm$0.58 &2.05$\pm$1.10 &1.63 (\textcolor[RGB]{254,76,97}{-0.23}) & 1.49 (\textcolor[RGB]{254,76,97}{-0.21}) \\ 
Joint V1 (w/ CR) & 1.29$\pm$0.26&  1.65$\pm$0.44& 1.28$\pm$0.30& 1.13$\pm$0.28& 1.34 (\textcolor[RGB]{254,76,97}{-0.20})& 1.76$\pm$0.87& \textbf{1.15$\pm$0.36} &1.48$\pm$0.52 &\textbf{1.92$\pm$1.18} &1.58 (\textcolor[RGB]{254,76,97}{-0.28}) & 1.46 (\textcolor[RGB]{254,76,97}{-0.24}) \\ 
Joint V2 (w/ CR) & \textbf{1.27$\pm$0.28}&  1.73$\pm$0.47& 1.34$\pm$0.32& 1.08$\pm$0.25& 1.36 (\textcolor[RGB]{254,76,97}{-0.18})& 1.69$\pm$0.85& 1.18$\pm$0.47 &1.50$\pm$0.64 &1.97$\pm$1.26 &1.59 (\textcolor[RGB]{254,76,97}{-0.27}) & 1.47 (\textcolor[RGB]{254,76,97}{-0.23}) \\ 

\textbf{Ours (w/ CR)} & 1.31$\pm$0.27&  \textbf{1.49$\pm$0.38}& \textbf{1.22$\pm$0.27}& \textbf{1.00$\pm$0.24}& \textbf{1.26} (\textcolor[RGB]{254,76,97}{-0.28})& \textbf{1.55$\pm$0.78}& 1.24$\pm$0.34 &\textbf{1.27$\pm$0.32} & 2.01$\pm$0.95 &\textbf{1.52} (\textcolor[RGB]{254,76,97}{-0.34}) & \textbf{1.39} (\textcolor[RGB]{254,76,97}{-0.31}) \\ 
\hline
 \end{tabular}}

  \caption{The performance of cardiac substructure segmentation by using 2D Transformer.
  %by \textcolor[RGB]{0 205 102}{green}.
  }
 \label{tbl:cardiac}
 \vspace{-0.4cm}
\end{table*}
\subsubsection{Abdominal Multi-organ Segmentation.}
Likewise, the multi-modal Transformer architecture {Joint V1} improves the Dice value by $0.4\%$ on CT and $0.2\%$ on MRI when compared to the {Baseline} model trained from a single modality. 
%
%This indicates that Transformer has the advantage of analyzing multi-modal data even in 3D. 
%
In contrast to {Joint V1}, sharing decoder in {Joint V2} causes a slight segmentation performance drop when the number of parameters is reduced.
By integrating modality-specific activation, {Joint V3} improves the Dice values to $91.4\%$ on CT and $91.6 \%$ on MRI, outperforming both {Joint V2} and {Joint V1} by a significant margin. 
Furthermore, by using the $\mathcal{L}_{\rm{mcr}}$ and $\mathcal{L}_{\rm{icr}}$, all three models improve significantly, 
and our full scheme achieves the best segmentation results of $93.3\%$ overall mean Dice and $0.76 \rm{mm}$ overall mean ASD. Finally, we exhibit the visual segmentation results for quantitative comparison, as shown in Fig.~\ref{fig:cardiac_abdominal} (Right).

\subsection{Effectiveness of Each Key Component}

We employ four settings to verify the contribution of various key components:
(a) we train the {Joint V3} model without using any consistency regularization terms for both CT and MRI; 
(b) we only add the modality-level consistency regularization $\mathcal{L}_{\rm{mcr}}$ onto {Joint V3}, which corresponds to Eqn.10.; 
(c) we only add the instance-level consistency regularization $\mathcal{L}_{\rm{icr}}$ onto {Joint V3}, which corresponds to Eqn.11.; 
(d) we add both $\mathcal{L}_{\rm{mcr}}$ and $\mathcal{L}_{\rm{icr}}$ to accomplish our multi-modal learning scheme.

Table~\ref{tbl:ablation} reports the mean value of Dice and ASD for each class. 
Adding $\mathcal{L}_{\rm{mcr}}$ to {Joint V3} improves the average Dice to $92.9\%$ on CT and $92.2\%$ on MRI and decreases the average ASD to $0.83\rm{mm}$ on CT and $0.94\rm{mm}$ on MRI. 
We also observe that the segmentation performance of each class improves significantly, regardless of CT or MRI, proving that aligning the representations of each class across various modalities could narrow the modalities' discrepancies in data distribution, allowing the network to be more generalized for both types of data.
% shortening the distance between two modality-aware queries could explicitly align category representations of two modalities, making the global promotion for all categories. 
%
Furthermore, only adding $\mathcal{L}_{\rm{icr}}$ onto {Joint V3} results in a mean Dice improvement of $1.8\%$ on CT and $0.7\%$ on MRI. 
This demonstrates that facilitating the network to dynamically learn the consistency of inter-class relationships at the image level within different modalities could also enhance the network's generalization to different modalities.
% alignment of CT and MRI's category relationships at the instance level help the model to learn cross-modality semantic association and consistency. 
%
Finally, by including both $\mathcal{L}_{\rm{mcr}}$ and $\mathcal{L}_{\rm{icr}}$, the Dice value further improves to $93.7\%$ on CT and $92.8\%$ on MRI, 
outperforming the variants that only add $\mathcal{L}_{\rm{mcr}}$ and $\mathcal{L}_{\rm{icr}}$, 
verifying that the two regularization terms can be used jointly to pursue the structured semantic consistency and effectively improve the segmentation performance. 

\begin{figure*}[ht]
	\begin{center}
		\includegraphics[width=0.6\textwidth]{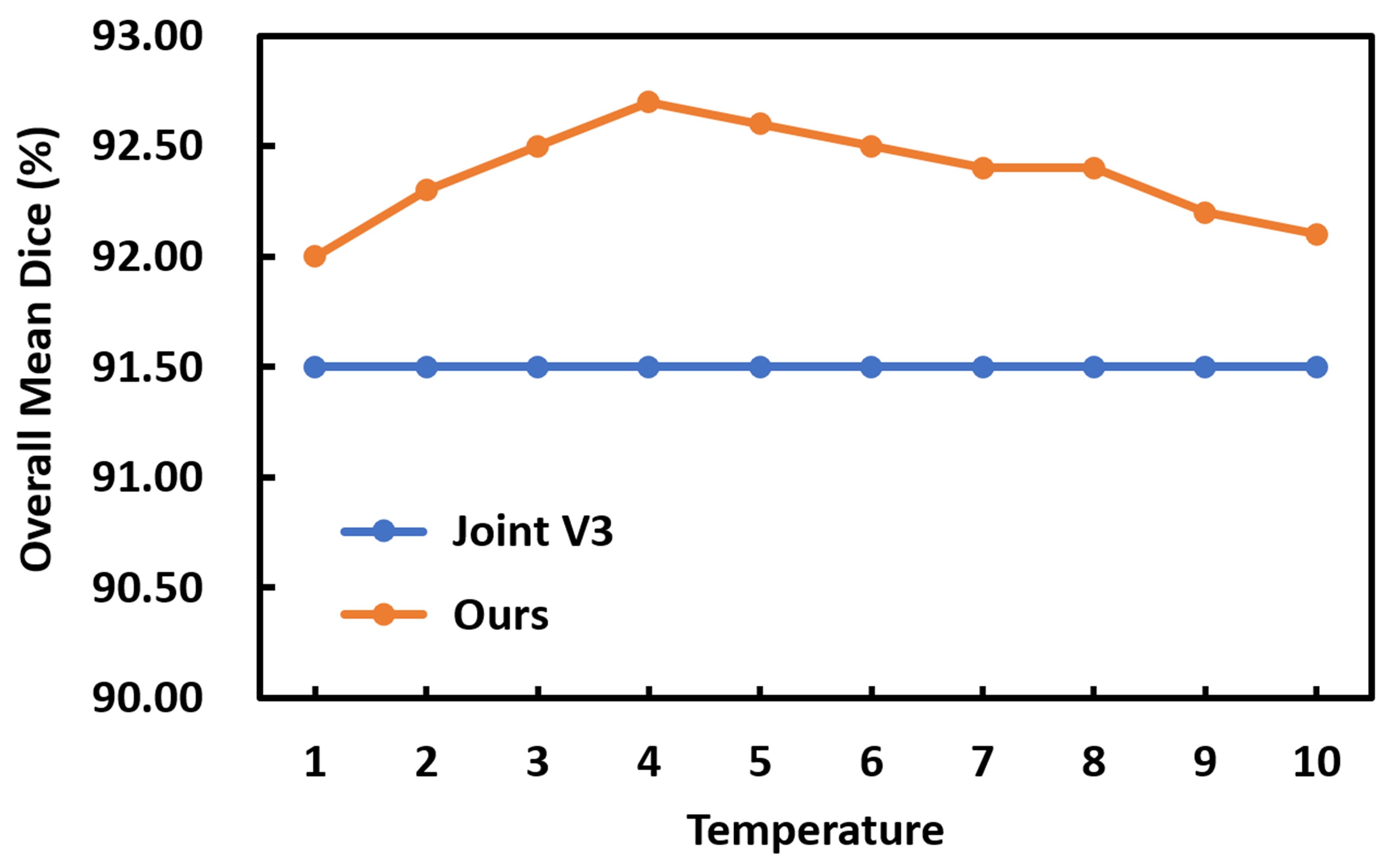}
	\end{center}
	\vspace{-10pt}	\caption{The effect of temperature $\tau$. Joint V3 is only used for comparison.} 
	\label{fig:temperature}
%\vspace{-0.5cm}
\end{figure*}

\subsection{Setting of Temperature Hyper-parameters}

The hyperparameter $\tau$ in $\mathcal{L}_{\rm{icr}}$ is used to control the softness of the inter-class probability distributions.
We vary the value of $\tau$ to see how it affects the final segmentation results.
As $\tau$ only influences the $\mathcal{L}_{\rm{icr}}$, we implement the {Joint V3} model with only $\mathcal{L}_{\rm{icr}}$ added. 
Fig.~\ref{fig:temperature} presents the segmentation performance changes (\textit{i.e.} overall mean Dice). 
As the temperature $\tau$ increases from $1$ to $4$, the inter-class relation retains richer semantic information to guide the cross-modality learning. 
However, when $\tau>4$, the more difficult pixel-wise constraints are exploited, and segmentation performance begins to deteriorate, possibly due to optimization difficulties.
Based on the above results, we use $\tau=4$ in all subsequent experiments.

\begin{table*}
\centering
\resizebox{\textwidth}{!}{
\begin{tabular}{l|ccccl|ccccl|l} 
\hline
\toprule [1.5 pt]
\multirow{2}{*}{Methods} & \multicolumn{5}{c|}{Abdominal CT} &
\multicolumn{5}{c|}{Abdominal MRI} & \multirow{2}{*}{Overall Mean}\\
\cline{2-11} 
   &  Liver & Spleen & R-kdy & L-kdy & Mean &  Liver & Spleen & R-kdy & L-kdy & Mean & \\ \specialrule{0.05em}{0pt}{3pt}
\multicolumn{12}{c}{Dice Coefficient (avg.$\pm$ std., $\%$) $\uparrow$}  \\ \specialrule{0.05em}{3pt}{0pt} 
Backbone  & 93.2$\pm$2.8 & 90.9$\pm$2.4&  86.9$\pm$2.7 & 87.5$\pm$3.8 & 89.5& 91.7$\pm$4.0 & 87.2$\pm$3.2 &90.9$\pm$2.7 &90.6$\pm$3.3 &90.1 & 89.9 \\  
Baseline & 93.6$\pm$2.2&  92.1$\pm$2.6 & 87.9$\pm$1.8 & 87.7$\pm$3.6 & 90.3 & 92.9$\pm$3.3& 87.8$\pm$2.9 &92.0$\pm$2.5 &91.4$\pm$3.1 &91.0 & 90.7 \\  \hline
Joint V1 & 93.9$\pm$2.5 &  92.7$\pm$2.3& 88.2$\pm$2.0 & 88.1$\pm$3.2 & 90.7 (\textcolor[RGB]{254,76,97}{+0.4})& 93.2$\pm$2.8 & 88.1$\pm$2.5  &92.4$\pm$2.3 &91.1$\pm$3.3 &91.2 (\textcolor[RGB]{254,76,97}{+0.2}) & 91.0 (\textcolor[RGB]{254,76,97}{+0.3}) \\ 
Joint V2 & 94.0$\pm$2.6 &  92.5$\pm$2.2 & 87.8$\pm$2.9 & 87.9$\pm$3.4 & 90.6 (\textcolor[RGB]{254,76,97}{+0.3})& 92.6$\pm$3.4 & 87.3$\pm$3.1 & 91.2$\pm$1.9 &90.8$\pm$3.7 &90.5 (\textcolor[RGB]{0 205 102}{-0.5})& 90.5 (\textcolor[RGB]{0 205 102}{-0.2}) \\ 
Joint V3 & 94.6$\pm$2.2 &  93.3$\pm$1.8 & 88.9$\pm$2.3 & 88.7$\pm$3.5& 91.4 (\textcolor[RGB]{254,76,97}{+1.1})& 93.8$\pm$2.3 & 88.5$\pm$2.7 &92.7$\pm$1.5 &91.5$\pm$3.1 &91.6 (\textcolor[RGB]{254,76,97}{+0.6}) & 91.5 (\textcolor[RGB]{254,76,97}{+0.8}) \\ 
Joint V1 (w/ CR) & 95.3$\pm$1.8&  94.3$\pm$1.1 & 91.6$\pm$1.0 & 91.2$\pm$2.5 & 93.1 (\textcolor[RGB]{254,76,97}{+2.8})& 94.2$\pm$1.7& 89.3$\pm$2.1 &93.3$\pm$1.1 &92.6$\pm$2.3 &92.4 (\textcolor[RGB]{254,76,97}{+1.4}) & 92.7 (\textcolor[RGB]{254,76,97}{+2.0}) \\ 
Joint V2 (w/ CR) & 95.1$\pm$1.7 & 93.9$\pm$1.5 & 91.3$\pm$1.4& 91.0$\pm$2.7& 92.8 (\textcolor[RGB]{254,76,97}{+2.5})& 94.0$\pm$1.9& 89.1$\pm$1.8 &92.8$\pm$0.6 &92.2$\pm$2.0 &92.0 (\textcolor[RGB]{254,76,97}{+1.0}) & 92.4 (\textcolor[RGB]{254,76,97}{+1.7}) \\ 
\textbf{Ours (w/ CR)} & \textbf{95.8$\pm$1.4} &  \textbf{94.9$\pm$1.3} & \textbf{92.3$\pm$0.9} & \textbf{91.8$\pm$2.2} & \textbf{93.7} (\textcolor[RGB]{254,76,97}{+3.4})& \textbf{94.7$\pm$1.5} &\textbf{89.9$\pm$1.2} &\textbf{93.6$\pm$0.8}& \textbf{93.0$\pm$1.4} &\textbf{92.8} (\textcolor[RGB]{254,76,97}{+1.8}) & \textbf{93.3} (\textcolor[RGB]{254,76,97}{+2.6})  \\ \specialrule{0.05em}{0pt}{3pt}
\multicolumn{12}{c}{Average Symmetric Surface
Distance (avg.$\pm$ std., $mm$) $\downarrow$}  \\ \specialrule{0.05em}{3pt}{0pt} 
Backbone & 1.19$\pm$0.91 &   1.18$\pm$0.82&1.84$\pm$1.06 & 1.10$\pm$0.78 & 1.33& 1.20$\pm$0.68& 1.27$\pm$0.79 &1.36$\pm$0.94 &1.37$\pm$0.71 &1.30 & 1.31 \\ 
Baseline & 1.12$\pm$0.75&  0.98$\pm$0.68& 1.60$\pm$0.93&  1.05$\pm$0.65& 1.19& 1.07$\pm$0.52& 1.19$\pm$0.76 &1.22$\pm$0.80 &1.23$\pm$0.64 &1.18 & 1.18 \\ \hline
Joint V1 & 1.07$\pm$0.56&  0.79$\pm$0.45 & 1.45$\pm$0.71 & 0.93$\pm$0.49& 1.06 (\textcolor[RGB]{254,76,97}{-0.13})& 1.12$\pm$0.46 & 1.06$\pm$0.52 &1.18$\pm$0.62 & 1.30$\pm$0.55&1.17 (\textcolor[RGB]{254,76,97}{-0.01}) & 1.11 (\textcolor[RGB]{254,76,97}{-0.07}) \\ 
Joint V2 & 1.03$\pm$0.62&  0.85$\pm$0.51& 1.87$\pm$0.84 & 0.96$\pm$0.58& 1.18 (\textcolor[RGB]{254,76,97}{-0.01})& 1.19$\pm$0.56 & 1.32$\pm$0.73 &1.27$\pm$0.85 &1.34$\pm$0.67 &1.28 (\textcolor[RGB]{0 205 102}{+0.10}) & 1.23 (\textcolor[RGB]{0 205 102}{+0.05}) \\ 
Joint V3 & 0.94$\pm$0.58 &  0.75$\pm$0.37 & 1.37$\pm$0.61 & 0.82$\pm$0.43 & 0.97 (\textcolor[RGB]{254,76,97}{-0.22})& 1.01$\pm$0.49& 1.18$\pm$0.64 &1.03$\pm$0.69 &1.15$\pm$0.53 &1.09 (\textcolor[RGB]{254,76,97}{-0.09}) & 1.03 (\textcolor[RGB]{254,76,97}{-0.15}) \\ 
Joint V1 (w/ CR) & \textbf {0.72$\pm$0.25} &  0.68$\pm$0.23 & 0.95$\pm$0.37 & \textbf {0.70$\pm$0.18} & 0.76 (\textcolor[RGB]{254,76,97}{-0.43})& 0.91$\pm$0.37 & 0.86$\pm$0.48 &0.94$\pm$0.43 &0.88$\pm$0.32 &0.90 (\textcolor[RGB]{254,76,97}{-0.28}) & 0.83 (\textcolor[RGB]{254,76,97}{-0.35}) \\ 
Joint V2 (w/ CR) & 0.91$\pm$0.41 &  0.72$\pm$0.29& 1.06$\pm$0.43 & 0.75$\pm$0.22 & 0.86 (\textcolor[RGB]{254,76,97}{-0.33})& 0.89$\pm$0.30 & 0.92$\pm$0.57 &0.97$\pm$0.56 &0.94$\pm$0.42 &0.93 (\textcolor[RGB]{254,76,97}{-0.25}) & 0.90 (\textcolor[RGB]{254,76,97}{-0.28}) \\ 
\textbf{Ours (w/ CR)}  & 0.87$\pm$0.29 &  \textbf {0.58$\pm$0.17} & \textbf {0.84$\pm$0.32} & 0.72$\pm$0.24 & \textbf {0.75} (\textcolor[RGB]{254,76,97}{-0.44}) & \textbf {0.83$\pm$0.36}& \textbf {0.56$\pm$0.23} & \textbf {0.85$\pm$0.39} &\textbf {0.83$\pm$0.37} & \textbf {0.77} (\textcolor[RGB]{254,76,97}{-0.41}) & \textbf {0.76} (\textcolor[RGB]{254,76,97}{-0.42})  \\
\hline
 \end{tabular}}
  \caption{The performance of abdominal multi-organ segmentation by using 3D Transformer.}
 \label{tbl:organ}
   %\vspace{-0.5cm}
\end{table*}

\begin{table*}
\centering
\resizebox{\textwidth}{!}{
\begin{tabular}{l|ccccl|ccccl|l} 
\toprule [1.5 pt]
\multirow{2}{*}{Methods}& \multicolumn{5}{c|}{Abdominal CT} &
\multicolumn{5}{c|}{Abdominal MRI} & \multirow{2}{*}{Overall Mean} \\
\cline{2-11} 
   &  Liver & Spleen & R-kdy & L-kdy & Mean &  Liver & Spleen & R-kdy & L-kdy & Mean & \\ \specialrule{0.05em}{0pt}{3pt}
\multicolumn{12}{c}{Dice Coefficient (avg.$\pm$ std., $\%$) $\uparrow$}  \\ \specialrule{0.05em}{3pt}{0pt}
Joint V3 & 94.6$\pm$2.2 &  93.3$\pm$1.8 & 88.9$\pm$2.3 & 88.7$\pm$3.5& 91.4& 93.8$\pm$2.3 & 88.5$\pm$2.7 &92.7$\pm$1.5 &91.5$\pm$3.1 &91.6 & 91.5 \\ 
Joint V3 + $\mathcal{L}_{\rm{mcr}}$ & 95.3$\pm$2.0&  94.4$\pm$1.9  & 91.1$\pm$1.5& 90.8$\pm$3.1 & 92.9 (\textcolor[RGB]{254,76,97}{+1.5}) & 94.2$\pm$1.8 & 89.0$\pm$1.7 &93.3$\pm$1.2 &92.1$\pm$1.9 &92.2 (\textcolor[RGB]{254,76,97}{+0.6}) & 92.5 (\textcolor[RGB]{254,76,97}{+1.0}) \\

Joint V3 + $\mathcal{L}_{\rm{icr}}$ & 95.1$\pm$1.8 & 94.5$\pm$1.6 & 91.6$\pm$1.7& 91.6$\pm$2.8 & 93.2 (\textcolor[RGB]{254,76,97}{+1.8}) & 94.4$\pm$1.6 & 89.3$\pm$1.5 &93.0$\pm$1.6 &92.4$\pm$2.2 & 92.3 (\textcolor[RGB]{254,76,97}{+0.7}) & 92.7 (\textcolor[RGB]{254,76,97}{+1.2}) \\ 

Joint V3 + $\mathcal{L}_{\rm{mcr}}$ + $\mathcal{L}_{\rm{icr}}$ & 95.8$\pm$1.4 &  94.9$\pm$1.3 & 92.3$\pm$0.9 & 91.8$\pm$2.2 & 93.7 (\textcolor[RGB]{254,76,97}{+2.3})& 94.7$\pm$1.5 &89.9$\pm$1.2 &93.6$\pm$0.8& 93.0$\pm$1.4 &92.8 (\textcolor[RGB]{254,76,97}{+1.2}) & 93.3 (\textcolor[RGB]{254,76,97}{+1.8}) \\ \specialrule{0.05em}{0pt}{3pt}
\multicolumn{12}{c}{Average Symmetric Surface
Distance (avg.$\pm$ std., $mm$) $\downarrow$}  \\ \specialrule{0.05em}{3pt}{0pt} 
Joint V3 & 0.94$\pm$0.58 &  0.75$\pm$0.37 & 1.37$\pm$0.61 & 0.82$\pm$0.43 & 0.97& 1.01$\pm$0.49& 1.18$\pm$0.64 &1.03$\pm$0.69 &1.15$\pm$0.53 &1.09 & 1.03 \\ 
Joint V3 + $\mathcal{L}_{\rm{mcr}}$ & 0.91$\pm$0.46 & 0.67$\pm$0.31  & 0.98$\pm$0.52 & 0.77$\pm$0.37  &0.83 (\textcolor[RGB]{254,76,97}{-0.14}) & 0.96$\pm$0.43 & 0.79$\pm$0.45  &0.98$\pm$0.66 &1.04$\pm$0.49 &0.94 (\textcolor[RGB]{254,76,97}{-0.15}) & 0.89 (\textcolor[RGB]{254,76,97}{-0.14})\\ 
Joint V3 + $\mathcal{L}_{\rm{icr}}$ & 0.88$\pm$0.37 & 0.65$\pm$0.28 & 0.93$\pm$0.43& 0.74$\pm$0.33& 0.80 (\textcolor[RGB]{254,76,97}{-0.17})& 0.91$\pm$0.35& 0.74$\pm$0.38 &0.89$\pm$0.51 &0.96$\pm$0.45 &0.88 (\textcolor[RGB]{254,76,97}{-0.21}) & 0.84 (\textcolor[RGB]{254,76,97}{-0.19}) \\ 
Joint V3 + $\mathcal{L}_{\rm{mcr}}$ + $\mathcal{L}_{\rm{icr}}$ & 0.87$\pm$0.29 &  0.58$\pm$0.17 & 0.84$\pm$0.32 & 0.72$\pm$0.24 & 0.75 (\textcolor[RGB]{254,76,97}{-0.22})& 0.83$\pm$0.36& 0.56$\pm$0.23 & 0.85$\pm$0.39 &0.83$\pm$0.37 & 0.77 (\textcolor[RGB]{254,76,97}{-0.32})& 0.76 (\textcolor[RGB]{254,76,97}{-0.27}) \\
\hline
 \end{tabular}}
 \caption{Ablation studies on abdominal multi-organ segmentation with 3D Transformer.}
 \label{tbl:ablation}
 \vspace{-0.5cm}
\end{table*}

\section{Discussion}
This paper aims to address cross-modal medical image segmentation based on the unpaired training samples, \textit{e.g.,} CT and MRI images.
Such multi-modal learning allows a single model to analyze data from multiple imaging devices, which greatly improves the efficiency of data usage.
%
%Specifically, we can leverage cross-modal information to make better use of the limited annotated samples and achieve mutual promotion of the performance of each modality. In practice, the main difficulty is caused by the distinct physical principles of the underlying image acquisition, leading to the large visual variance of multi-modal images. There exists limited studies to address such an open issue. In the literature, Dou \textit{et al.} \cite{dou2020unpaired} proposed a pioneer work to reuse most of network parameters, by sharing all convolutional kernels across different modalities, and conduct specific modality predictions by switching modality-specific internal normalization layers. This approach is more about learning to store idiosyncratic information of certain modalities in the corresponding normalization layers, rather than embedding semantic associations between modalities into the network parameters through model training. Moreover, such a method requires maintaining different normalization layers for different modalities, which is unfriendly during the deployment of the model, especially for embedded intelligent devices.
%
%
To better exploit the cross-modality information, we propose a novel method to accomplish cross-modal segmentation through learning structured semantic consistency between different modalities.
Our model is designed for unpaired multi-modal images and is stable during the training phase.
To learn structured semantic consistency across modalities (\textit{i.e. the consistencies of semantic class representations and their
correlations.}), 
we introduce a carefully designed External Attention Module (EAM) to conduct semantic consistency regularizations both at the modality and image levels.
Such a module is very simple and flexible to use.
It is only an external module used to embed cross-modal semantic consistency into the backbone network during the training phase, thus can be removed during the testing phase, ensuring the simplicity of the model.
Moreover, the input of EAM is only the feature maps extracted at the specific scale. Therefore, it can be easily integrated onto various existing 2D and 3D Transformer architectures.

During the training process, we first construct globally learnable class embeddings for each modality, with the goal of capturing the representation of each class within each modality. Given that we use the same label taxonomy for unpaired CT and MRI, one intuitive strategy for learning consistent semantic information is to directly align class representations across modalities.
% we encourage aligning the representations of each class across various modalities since the same classes exist in our setting for the different unpaired modalities. 
However, we find that globally aligned class representations will not render the network more robust to sample variations. Driven by such a discovery, we further encourage the network to learn consistent semantic information at the image-level.
% perform semantic alignment at the image level. 
%
The previous approach \cite{dou2020unpaired} directly align confusion matrices of predicted results across modalities. In contrast, 
we highlight semantic propagation at multiple scales from global to local, by interacting the global class representations across the entire dataset with the semantic features of each image, so as to learn image-specific class representations. Then, we derive multi-scale inter-class correlations within each image and dynamically establish its consistency between different modal data during training. 

We conduct extensive evaluations on two medical image segmentation scenarios, outperforming the state-of-the-art methods with a large margin, \textit{i.e.} 2.6$\%$ and 2.5$\%$ improvements on overall mean Dice for two tasks respectively. We further utilize few-shot setting to see how our method performs when one modality has far fewer samples than the other. Surprisingly, we find that a modality with a small number of training samples can boost the training of another modality with a large number of training samples, and a modality with a large number of training samples can greatly supplement the problem of another modality with a small number of training samples.
And it is worth noting that since Transformer requires a large amount of data for training, we still initialize the model pretrained on Image-Net \cite{deng2009imagenet} for both datasets due to the limited data availability, otherwise, the performance will suffer.

%We will continue to train on large-scale medical datasets in the future to validate our methods, such as MICCAI AMOS 2022 \cite{ji2022amos}.
% Talk about the findings through the experiment.

A few limitations of the proposed method should be mentioned. 
Although the proposed method outperforms the state-of-the-art unpaired multi-modal learning schemes, the segmentation accuracy  still has very huge room for improvement since we only use the basic Transformer architectures.
This should be acceptable since we mainly focus on learning to align the structured semantic information from different modalities but not the detailed backbone network architecture design for pixel-level predictions.
In addition, although the proposed method allows exploring more modalities (\textit{e.g.,} CT, MRI, and X-ray) to learn semantic consistencies simultaneously.
When the number of modalities increases, how to align the semantic consistencies across these modalities is still not well addressed.
One of the most straightforward ways is to group all modalities in pairs and align them one group by another.
However, such a scheme may not the most efficient one when training the model. 
We plan to design a simple yet effective strategy to tackle the above issue in our future work.

Overall, we propose a novel scheme to learn structured semantic consistency between different modalities from unpaired samples via an attention mechanism.
We apply it to the joint semantic segmentation of CT and MRI, whose appearances have a large discrepancy, and it achieves significant progress compared with counterparts.
Intuitively, such a scheme can be easily extended to other domain alignment problems.
For example, it can also learn unified abdominal organ representations from multiple datasets with different label taxonomy.
This is common in practice, \textit{e.g.,} some datasets label the \texttt{left kidney} and \texttt{right kidneys} as the same \texttt{kind}, while others label them as different semantic classes, or some datasets label the \texttt{intestines} as a single class, while others distinguish it with different segments.
To tackle such an issue, we just need to introduce a learnable transformation matrix in our proposed EAM module to learn the mapping relationships between different semantic labels. 
We will explore these extensions in future work.

\end{document}